\documentclass{article}
\usepackage[preprint]{neurips_2026}

\usepackage[utf8]{inputenc} 
\usepackage[T1]{fontenc}    
\usepackage{url}            
\usepackage{booktabs}       
\usepackage{amsfonts}       
\usepackage{nicefrac}       
\usepackage{microtype}      
\usepackage{amsmath}
\usepackage{amssymb}
\usepackage{amsthm}
\usepackage{xcolor}
\usepackage{graphicx}  
\usepackage{adjustbox}
\usepackage{subcaption}
\usepackage{tikz}
\usepackage{booktabs}
\usepackage{multirow}
\usepackage{geometry}
\usetikzlibrary{arrows.meta, positioning, fit, calc,
                backgrounds, decorations.pathreplacing}
\definecolor{linkblue}{rgb}{0.1, 0.45, 0.8}
\usepackage[colorlinks=true, citecolor=linkblue, linkcolor=linkblue, urlcolor=linkblue]{hyperref}       
\usepackage[capitalize]{cleveref}   
\usepackage{enumitem}
\theoremstyle{plain}
\newtheorem{theorem}{Theorem}[section]

\theoremstyle{definition}
\newtheorem{definition}[theorem]{Definition}

\theoremstyle{remark}

\title{A Critical Audit of Spatiotemporal Forecasting Benchmark Datasets and Models}

\author{%
  Kenneth Martin\thanks{Equal contribution.}\\
  Imperial College London\\
  London, United Kingdom\\
  \texttt{k.martin24@imperial.ac.uk}\And
  Simon Heilig\footnotemark[1]\\
  Ruhr University Bochum\\
  Bochum, Germany\\
  \texttt{simon.heilig@rub.de}\And
  Asja Fischer\\
  Ruhr University Bochum\\
  Bochum, Germany\\
  \And
  Michel F.\,C. Haddad\\
  Queen Mary University of London\\
  London, United Kingdom
  \And
  Adam M. Sykulski\\
  Imperial College London\\
  London, United Kingdom
  \And
  Moshe Eliasof \\
  Ben-Gurion University of the Negev\\
  Beersheba, Israel
}

\begin{document}

\maketitle

\begin{abstract}
Graph neural networks (GNNs) are routinely employed for spatiotemporal forecasting, yet their performance across widely used benchmark datasets is inconsistent. Here, we perform an audit of dataset properties and baseline models to assess the quality of the benchmarks, and the robustness of the conclusions drawn from them. Using classical statistical tools, we characterise spatiotemporal lagged dependencies in benchmarks, and examine how temporal differencing changes these relationships and affects model rankings. Motivated by this, we re-evaluate temporal linear baselines, significantly reducing the apparent gains from GNNs on several benchmarks, and surpassing GNNs on others. Suspecting that GNNs struggle to extract linear, node-wise signals, we find that supplying them with autoregressive residuals improves their performance particularly on non-traffic benchmarks. Finally, controlled synthetic experiments reveal that GNNs are sensitive to heterogeneity in temporal dynamics and spatial graph interactions. Together, our findings demonstrate that baseline specification, data pre-processing and system heterogeneity shape the interpretations drawn from benchmark rankings, informing the design and robust evaluation of GNNs.
\end{abstract}

\section{Introduction}
Graph neural networks (GNNs) have seen great promise in applications requiring spatial information processing \citep{shen2022graph,corso2024graph,du2024densegnn} due to their inductive bias stemming from the message passing mechanism, aggregating information over irregular spatial measurements. With the advancement of filters for signals defined on a graph \citep{defferrard2016convolutional,kipf2017semisupervised,graphsignalproc}, generalizing the convolution neural network (CNN) beyond grid-structured data, and more recently the development of neural ODE-inspired GNNs \citep{poli2019graph,pmlr-v202-choi23a}, a natural link to processing data spread across both space and time has emerged \citep{li2018diffusion,STGCN,li2024amgcn,eliasof2024temporal,eliasof2025graph,ceni2025message}. The common taxonomy differentiates between static graphs, i.e., fixed spatial structure, and dynamic graphs, i.e., spatial interactions that vary in time \citep{gravina2024deep}. In this paper we focus on static graphs equipped with time series features, representing a multivariate forecasting problem based on sparse spatial interactions given by a graph defined in the data scenario, e.g., traffic congestion or disease forecasting \citep{corradini2025systematic}. Accurate forecasting in these settings requires models to represent local temporal dynamics, which may differ amongst nodes, and predictive interactions between nodes. It is this variation in local temporal dynamics that we refer to as temporal heterogeneity.

Reliable evaluation, however, requires appropriate baselines and consistent forecasting protocols. Datasets such as Chickenpox, PedalMe, WikiMaths, METR-LA, PEMS-BAY and PEMS03/04/07/08 \citep{rozemberczki2021pytorch,li2018diffusion,song2020spatial}, which are distributed through widely used libraries, have become de facto standard benchmarks for spatiotemporal forecasting. Coming from the model perspective, simple but strong forecasting baselines have been repeatedly demonstrated in the past: \citet{MICHELI202285} evaluated constant predictors and node-wise ridge regression, demonstrating competitive performance on Chickenpox, PedalMe, and WikiMaths. DLinear \citep{zeng2023transformers} demonstrated the competitiveness of linear forecasters in long term-forecasting, particularly under extended input windows. STID \citep{shao2022stid} demonstrated that an MLP equipped with learned node, time-of-day and day-of-week embeddings can achieve competitive forecasting performance without graph message passing.

Considering the widespread use of these datasets, we perform a critical audit to examine how modelling choices, dataset properties and evaluation protocols shape model comparisons. However, effectively assessing improvements in forecasting architectures requires well-specified and well-tuned temporal models. We therefore re-evaluate linear and non-linear models across a suite of benchmark datasets, helping to establish how much performance neural graph architectures can add beyond local temporal modelling.
Alongside re-evaluating baseline models, one of the central questions of our paper is how temporal heterogeneity within the data challenges the models, and whether explicit node-wise modelling helps resolve possible difficulties. To this end, we make the following contributions:

\begin{itemize}[leftmargin=2em]
    \item A spatiotemporal lagged dependence analysis of widely adopted benchmark datasets: Chickenpox, PedalMe, WikiMaths, METR-LA, PEMS-BAY and PEMS03/04/07/08 (Section~\ref{sec:correlationana}); revealing a shortcoming that the official data-loader from the PyTorch Geometric Temporal library releases first-order differenced data for Chickenpox and PedalMe, which yields worse train-test generalization across the board when compared to the original (level) data (Section~\ref{sec:differencing}).
    \item Novel and simple empirical measures of temporal and spatial heterogeneity in spatiotemporal datasets, that provide explainability of subsequent forecasting performance (Section \ref{sub-sec:heterogeneity}).
    \item A protocol-matched re-evaluation of node-wise ARIMA models, showing that temporal-only models are competitive on Chickenpox and Wikimaths, and substantially narrow the gap on traffic datasets (METR-LA, PEMS-BAY and PEMS03/04/07/08). Subsequent integration of temporal-only models with GNNs demonstrate the added value of graph-based forecasting (Sections~\ref{sec:audit-protocols} and \ref{sec:audit_results}).
    \item A controlled synthetic experimental suite for examining the impact of temporal and spatial heterogeneity on forecasting performance; our analysis reveals that heterogeneity severely impacts performance across eleven different GNN models, and variants without a separate self-loop pathway fall behind when faced with lagged spatiotemporal responses (Section \ref{sec:synthetic}).
\end{itemize}

\section{Benchmark Datasets Audit}
The benchmark datasets and forecasting tasks are summarized in Appendix~\ref{app:experimental_setup}, Table~\ref{tab:datasets}. Before proceeding with our dataset audit, we first establish notation.

\paragraph{Notation.}
Let $\mathcal{G} = (\mathcal{V}, \mathcal{E}, \mathbf{A})$ be a weighted undirected graph with
$N = |\mathcal{V}|$ nodes and $|\mathcal{E}|$ edges, where $\mathbf{A} \in \mathbb{R}^{N \times N}$
is the adjacency matrix with $A_{ij} > 0$ if $(i,j) \in \mathcal{E}$ and zero otherwise.
Each node $i \in \mathcal{V}$ is associated with a scalar observation $y_{t}^{(i)} \in \mathbb{R}$
at discrete time $t \in \{1, \ldots, T\}$.
Stacking over nodes, we write the system state at time $t$ as $\mathbf{y}_t \in \mathbb{R}^N$,
and the full dataset as $\mathbf{Y} \in \mathbb{R}^{T \times N}$.

A forecasting task is defined as follows. Given a history window of length $H$,
i.e.\ $\mathbf{Y}_{t-H+1:t} \in \mathbb{R}^{H \times N}$, and the graph $\mathcal{G}$,
the goal is to predict the next $k$ steps
$\mathbf{Y}_{t+1:t+k} \in \mathbb{R}^{k \times N}$.
We denote the neighborhood of node $i$ as $\mathcal{N}(i) = \{j : (i,j) \in \mathcal{E}\}$,
and write the univariate time series for node $i$ as
$\mathbf{y}^{(i)} = (y^{(i)}_1, \ldots, y^{(i)}_T)^\top \in \mathbb{R}^T$.

\subsection{Spatiotemporal Lagged Dependence Analysis}\label{sec:correlationana}

\paragraph{Setup.} Based on established statistical tools \citep{shumway_time_2025}, we characterize temporal and graph-lagged linear dependence at up to three times the window length $H$. This allows us to examine dependence within and beyond canonical input windows, although longer seasonal patterns may lay outside.

By temporal correlation we refer to the autocorrelation of each
node's time series, i.e.\
\begin{equation}\label{eq:corrtemp}
    \rho^{\mathrm{temp}}_{i}(\tau)
    \;=\;
    \mathrm{Corr}\!\left( y^{(i)}_{t},\, y^{(i)}_{t-\tau} \right), \quad \tau = 0,1,2,\ldots.
\end{equation}
To examine dependence along the graph, we define the aggregated neighborhood signal
\begin{equation}\label{eq:neighbourhood}
    \bar y^{(i)}_{\tau} \;=\; \sum_{j\in\mathcal{N}(i)\setminus\{i\}} A_{ij}\, y^{(j)}_{\tau},
\end{equation}
with self-loops excluded. The graph-lagged spatial correlation is then given by
\begin{equation}
    \rho^{\mathrm{sp}}_{i}(\tau) \;=\; \mathrm{Corr}\!\left( y^{(i)}_{t},\, \bar y^{(i)}_{t-\tau} \right), \quad \tau = 0,1,2,\ldots.
\end{equation}

\paragraph{Partial spatial correlation.}
To assess the association remaining after accounting for temporal information available at the same lag, we project both the target and lagged neighborhood signal onto $y_{t-\tau}^{(i)}$. For each node, and lag $\tau$, let $\hat{y}^{(i)}_{t} = \alpha\, y^{(i)}_{t-\tau} + \beta$ and
$\hat{\bar{y}}^{(i)}_{t-\tau} = \gamma\, y^{(i)}_{t-\tau} + \delta$ be the
least-squares projections of $y^{(i)}_{t}$ and $\bar{y}^{(i)}_{t-\tau}$
onto $y^{(i)}_{t-\tau}$, respectively, where $\bar{y}^{(i)}_{t-\tau}$
is the aggregated neighborhood signal at lag $\tau$ as defined in~\eqref{eq:neighbourhood}.
The \textit{partial} spatial correlation is then
\begin{equation}\label{eq:psc}
    \rho^{\mathrm{psp}}_{i}(\tau)
    \;=\;
    \mathrm{Corr}\!\left(
        y^{(i)}_{t} - \hat{y}^{(i)}_{t},\;
        \bar{y}^{(i)}_{t-\tau} - \hat{\bar{y}}^{(i)}_{t-\tau}
    \right), \quad \tau = 1, 2, \ldots.
\end{equation}
\paragraph{Reporting and interpretation.}

In Appendix~\ref{appendix:statistics}, Figure~\ref{fig:appendix_acf} shows the descriptive profiles of our correlation functions defined above, over lags at up to three times the input window, for all datasets included in this paper. At each lag, we report the median, 5th and 95th percentile for each correlation function. The diagnostics we use only capture linear dependence at the examined lags, and only under the specific graph aggregation we use. Hence, this analysis does not take into account a possibly misspecified graph, multi-hop interactions, or non-linear relationships.

\paragraph{Findings.} 

The datasets exhibit contrasting dependence profiles. METR-LA, PEMS-BAY, PEMS03/04/07/08 show strong and consistent temporal and spatial correlations over the examined lags, with PEMS03/04/07/08 correlations decaying the slowest. Partial spatial correlation is relatively small, but remains strongly persistent for a large number of lags. WikiMaths exhibits a seasonal lag-7 pattern, which is exploitable in forecasting as the input window is 8 steps. The released version of Chickenpox (which first differences the data) shows a strong lag-1 negative temporal correlation, and weak partial spatial correlation, but the profiles are more similar to the traffic datasets (METR-LA, PEMS-BAY, PEMS03/04/07/08) when considering original, level data. PedalMe is also released as first differences but its short observation window means that dependence patterns are difficult to distinguish with the differenced or level data. However, a common observation amongst all datasets examined is that temporal autocorrelation profiles dominate spatial ones.

Although median profiles summarize typical dependence, even with our percentiles it can obscure the degree of variation amongst nodes. We quantify this variation as an empirical description of temporal and spatial heterogeneity in Section~\ref{sub-sec:heterogeneity}.

\subsection{Effects of differencing on Benchmark Representation.}\label{sec:differencing}
 As discussed in Section~\ref{sec:correlationana}, the distributed Chickenpox and PedalMe datasets from the PyTorch Geometric Temporal library contain first differences rather than original, level observations. Writing the level series as $Y_t$, the released representation is $Z_t = Y_t - Y_{t-1}.$ Differencing helps remove nonstationary components, but also materially changes the dependence that a forecasting model must represent \citep{hyndman_forecasting_2021}. In particular, differencing acts as a high-pass filter, attenuating low-frequency variation relative to higher-frequency components; an adapted proof of this from \citep{shumway_time_2025} can be found in Appendix~\ref{appendix:proof}. This motivates an examination of how model representation in level and first differences affects model ranking and forecasting performance.

\begin{figure}[t]
    \centering
    \begin{subfigure}[t]{0.49\linewidth}
        \centering
        \includegraphics[width=0.85\linewidth]{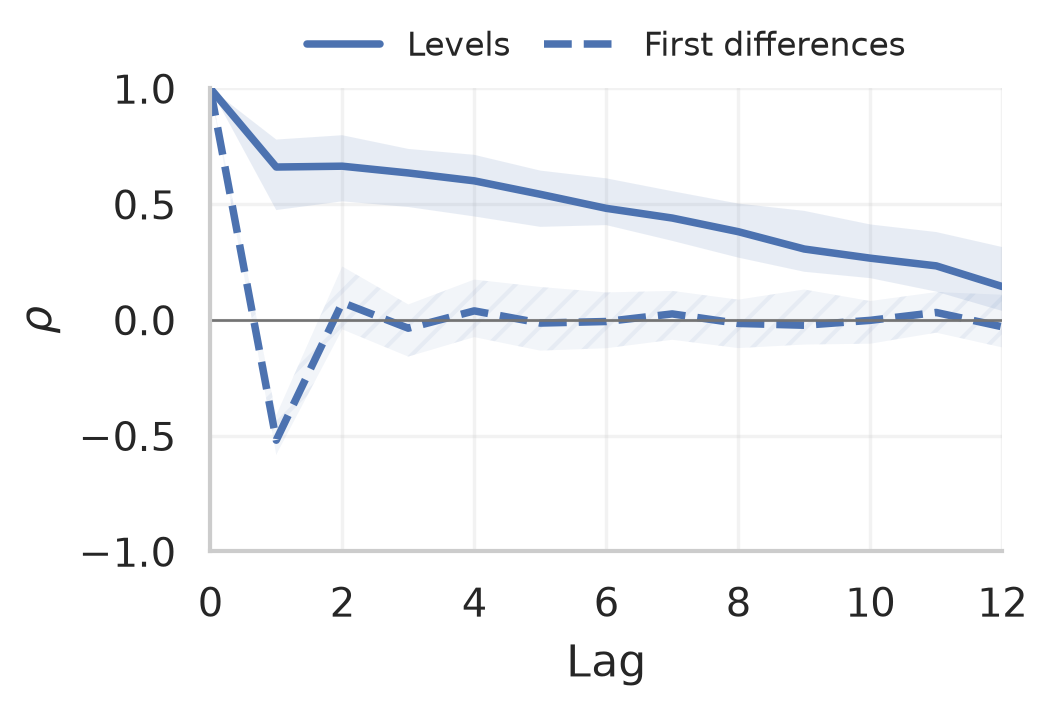}
        \caption{Chickenpox: temporal}
    \end{subfigure}\hfill
    \begin{subfigure}[t]{0.49\linewidth}
        \centering
        \includegraphics[width=0.85\linewidth]{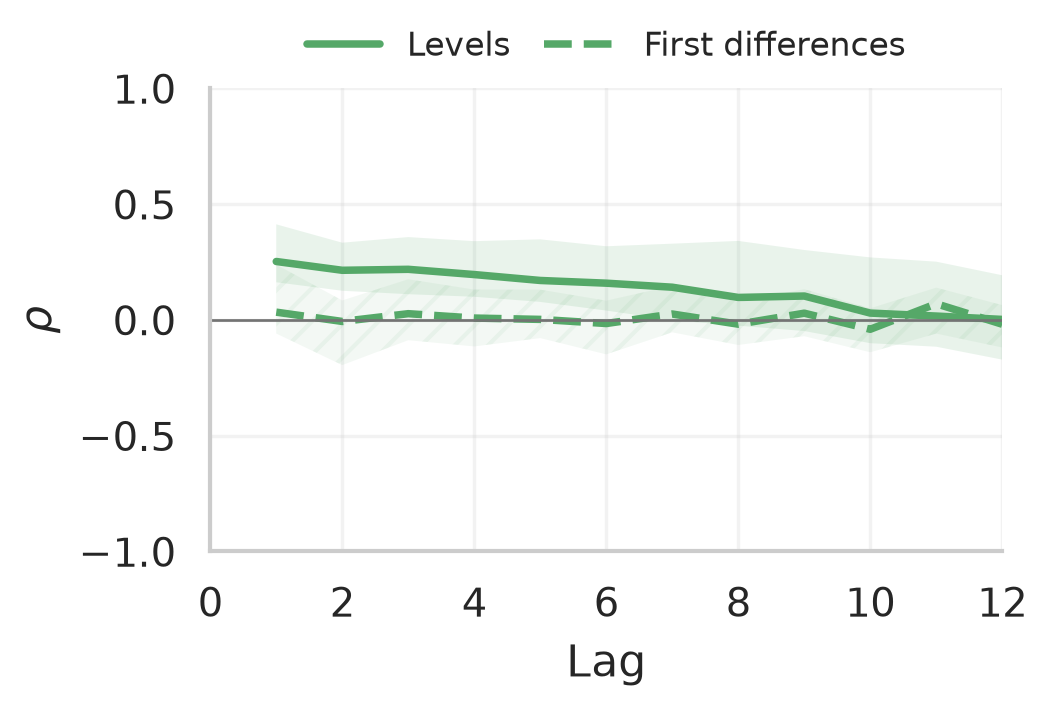}
        \caption{Chickenpox: partial spatial}
    \end{subfigure}
    \caption{Temporal (blue) and partial spatial (green) correlations
    for levels (solid) and first differences (dashed).
    Lines show nodewise medians; shaded regions span the 2.5th--97.5th
    percentiles across nodes, with hatching for first differences.
    These regions describe between-node variation, not confidence intervals.
    Lags extend to $3H=12$ for $H=4$.
    Partial spatial correlation controls for the same-lag own-node
    observation and is undefined at lag zero.}
    \label{fig:paired_corr}
\end{figure}

\paragraph{Changes in spatiotemporal lagged dependence structure.}
In Figure~\ref{fig:paired_corr} we display the temporal and partial spatial correlations of the first differenced and level datasets for Chickenpox. The level series exhibits strong temporal correlation up to 12 lags, whereas the first differences series has very weak temporal correlation left after only a few lags. Partial spatial correlation with Chickenpox is also more pronounced on the level dataset, with partial spatial signal visible up to 12 lags.
With PedalMe the differences are not as pronounced (see Figure~\ref{fig:appendix_acf}). PedalMe's temporal linear signal is slightly better recovered at lag 1, but the rest of the series has very little other structure. These differences show that pre-differencing materially affects the linear dependence available to forecasting models. To establish what kind of effect there is for non-linear signals, we look into comparisons with GNNs.

\paragraph{Changes in GNN performance.} Given these results, we hypothesized that GNNs trained on the first-differenced datasets would tend to do worse in generalization, compared to datasets trained on the level data. In Appendix~\ref{appendix:chickenpox}, Table~\ref{tab:iclr_differenced} we compare training and test errors for models fitted on the two representations on Chickenpox. The relative test-train gap, in percentages, is higher on differenced Chickenpox for 12 out of the 13 models we looked at, demonstrating a consistent change in generalization. A notable exception is TDE-GNN, which generalizes extremely well in both of standardized and level data already.

\subsection{Heterogeneity across nodes} \label{sub-sec:heterogeneity}
In Sections~\ref{sec:correlationana} and \ref{sec:differencing} we used our metrics to measure the overall strength of temporal and spatial dependence across nodes. This does not however establish how similarly this dependence behaves across nodes. We therefore adapt our correlation metrics to their natural extension as a measure of between-node variation in both temporal and partial spatial dependence.

\paragraph{Temporal and Spatial Heterogeneity Measures.}
Using the node-wise autocorrelation $\rho_i^{\mathrm{temp}}(\tau)$ introduced in \eqref{eq:corrtemp}, we can naturally define temporal dispersion across nodes over an input window of size $H$ as:
\begin{equation}
\mathcal{D}_T = \left[
        \frac{1}{HN}
        \sum_{\tau=1}^{H} 
        \sum_{i=1}^N
        \left(
            \rho_i^{\mathrm{temp}}(\tau) -
            \bar{\rho}_{\mathrm{temp}}(\tau)
        \right)^2
    \right]^\frac{1}{2},
\end{equation}
where $\bar{\rho}_{\mathrm{temp}}(\tau)$ is the average temporal correlation across all nodes at lag $\tau$.
This measure intuitively captures the dispersion of node autocorrelation functions across a window $H$. However, for a fixed level of dispersion, it is desirable for our measure to differentiate between systems with different overall strength of temporal signals. Therefore, we report the heterogeneity after accounting for the mean strength in dispersion. We call this quantity the temporal heterogeneity $\mathcal{H}_T$, where
\begin{equation}
\mathcal{H}_T=\frac{\mathcal{D}_T}{S_T}, \qquad 
S_T = \left[ 
    \frac{1}{HN}
    \sum_{\tau = 1}^H 
    \sum_{i=1}^N\rho_i^\mathrm{temp}(\tau)^2
    \right]^\frac{1}{2}.
\end{equation}

Similarly, we also define the partial-spatial graph heterogeneity $\mathcal{H}_G$ and dispersion $\mathcal{D}_G$ in the same way from~\eqref{eq:psc}. In Appendix~\ref{appendix:statistics}, Table~\ref{tab:heterogeneity} we display $\mathcal{H}_T$ and $\mathcal{H}_G$ for our different benchmarks with $H$ as given in Table~\ref{tab:datasets}. Full characterization over a larger range of lags can be found in Appendix~\ref{appendix:statistics}, Figure~\ref{fig:appendix_dispersion}.

\paragraph{Heterogeneity observations.}
Table~\ref{tab:heterogeneity} shows substantial differences amongst our examined datasets. Traffic datasets exhibit weak temporal heterogeneity: with the exception of METR-LA, all traffic datasets show $\mathcal{H}_T<0.1$. METR-LA has stronger temporal heterogeneity, but comparatively low partial spatial heterogeneity. For WikiMaths, examining its full profile in Figure~\ref{fig:appendix_dispersion}, we can observe that it has very high heterogeneity outside of its seasonal period (e.g., lags $1-6$), but for lag $7$ its signal becomes more homogeneous.

All benchmarks show very high partial-spatial graph heterogeneity, but particularly large for Chickenpox (differences).  If we examine Chickenpox (level), we can again observe a much more stable signal, with lower heterogeneity. PedalMe appears to be characterized by extremely high heterogeneity, but for its comparatively low number of nodes and timesteps, this is likely an artefact of noisy correlation estimates.

\paragraph{Modelling implications.}
These statistics characterize heterogeneity using observed linear temporal dependence. Hence, they cannot fully characterize heterogeneity in datasets. However, they do provide strong motivation for further analysis. In graph classification tasks, it has been shown that heterophilous datasets are more difficult for GNNs to analyze \citep{zhu_beyond_2020}. In the next section, we will test the analogous hypothesis for spatiotemporal forecasting by associating performance gaps between local temporal models and GNNs with our dispersion measures across benchmarks.

\section{A Model Audit}\label{sec:baselines}

The dependency analysis in Section \ref{sec:correlationana} suggests that much of the signal in all benchmarks is temporal, even if linearly. This raises the question: how much further improvement do GNNs provide, after that linear temporal signal is accounted for? Additionally, related to our analysis in Section \ref{sub-sec:heterogeneity}, since independently fitted models can also inherently adapt to heterogeneity naturally, they also form a good comparison for heterogeneity-related experiments. Both of these reasons justify the need for strong temporal baseline models.

Simple linear baselines, such as historical average and linear regression \citep{MICHELI202285} or seasonal averages and ARIMA (auto-regressive integrated moving average)~\citep{li2018diffusion}, have previously been used in the literature. However, when re-evaluating these linear models, we found that better specification and tuning substantially strengthened their performance relative to GNNs, as we now demonstrate.

\subsection{Evaluation Protocols}
\label{sec:audit-protocols}
\paragraph{Linear baseline models.}

Our node-wise baselines are autoregressive (AR) regression and ARIMA. On traffic benchmarks (METR-LA, PEMS-BAY and PEMS03/04/07/08), AR uses the previous twelve observations and fits a separate prediction for each forecast horizon, i.e. a 60-minute result is the horizon-12 prediction. 

Where datasets also have seasonal structure, we include linear model variants with these features embedded. On traffic datasets this is a practical protocol choice for fair comparisons, as canonically time-of-day features are given to GNNs. We add either 15-minute daily profile slots as coefficients, or embed it in the form of Fourier series. The Chickenpox variant of ARIMA+Fourier uses 6 yearly Fourier terms, where this is not strictly like-for-like with all GNNs, except TDE-GNN, which does use Fourier features. WikiMaths uses a seasonal ARIMA (SARIMA) with a 7-day period specification, which is contained in the input window. For further details on ARIMA models, see Appendix~\ref{sec:SARIMA}, and for dataset protocols, see Appendix~\ref{app:experimental_setup}. 

\paragraph{Non-linear models.}
 For traffic datasets, we use the 12-input 12-output benchmark structure associated with BasicTS+ \citep{shao2024exploring}. Our chosen models reported in the main text represent a selection of canonical models with established capabilities in the respective domain. For non-traffic datasets, no established protocol exists, so we compare with state-of-the-art model from \cite{eliasof2024temporal} and a traffic model from \cite{graphwavenet}.
\paragraph{Residual forecasts.}
If $\hat y^A_{t+h\mid t}$ is our first stage forecast, and $\hat e_{t+h\mid t}$ is the learned correction, the combined forecast is:
$$
\hat y_{t+h\mid t} = \hat y^A_{t+h\mid t} + \hat e_{t+h\mid t}.
$$
In practice, we provide each model with past observed first-stage model errors over its lookback window, while adding the linear model's next-step prediction to its MLP head. The non-traffic benchmarks use ARIMA models as their first-stage model, and GConvGRU~\citep{seo2018structured} and LSTM as spatial and non-spatial residual correctors. The traffic datasets use AR(12) models with added time-of-day embeddings, alongside Graph WaveNet~\citep{graphwavenet} and GConvGRU as the residual correctors.
 
 \subsection{Results}
 \label{sec:audit_results}
\begin{table*}[t]
\centering
\caption{Test MSE ($\downarrow$) under a chronological 9:1 train--test split. All results were run and independently confirmed by us. Neural and residual models are reported as mean $\pm$ standard deviation over 10 seeds; ARIMA models are deterministic. For WikiMaths, ARIMA-family models use a period-7 SARIMA specification. \textbf{Bold}, \underline{underline}, and \textit{italic} denote the first-, second-, and third-lowest MSE, respectively, within each dataset column; ties are retained.}
\label{tab:non-traffic-main}
\setlength{\tabcolsep}{6pt}
\begin{tabular}{lccc}
\toprule
Model & Chickenpox & PedalMe & WikiMaths \\
\midrule
\multicolumn{4}{l}{\textit{Neural and graph-based baselines}} \\
\addlinespace[2pt]
GConvGRU~\citep{seo2018structured}
  & 0.779\,{\scriptsize$\pm$0.009}
  & \underline{1.372\,{\scriptsize$\pm$0.005}}
  & \textit{0.443\,{\scriptsize$\pm$0.002}} \\
Graph WaveNet~\citep{graphwavenet}
  & 0.799\,{\scriptsize$\pm$0.017}
  & 1.424\,{\scriptsize$\pm$0.121}
  & \textbf{0.431\,{\scriptsize$\pm$0.005}} \\
TDE-GNN~\citep{eliasof2024temporal}
  & 0.782\,{\scriptsize$\pm$0.006}
  & \textbf{0.714\,{\scriptsize$\pm$0.051}}
  & 0.565\,{\scriptsize$\pm$0.017} \\
LSTM~\citep{hochreiter1997long}
  & 0.825\,{\scriptsize$\pm$0.005}
  & \textit{1.405\,{\scriptsize$\pm$0.026}}
  & 0.504\,{\scriptsize$\pm$0.002} \\
\midrule
\multicolumn{4}{l}{\textit{Statistical baselines and residual models}} \\
\addlinespace[2pt]
ARIMA
  & 0.827\,{\scriptsize$\pm$0.000}
  & 1.676\,{\scriptsize$\pm$0.000}
  & 0.479\,{\scriptsize$\pm$0.000} \\
ARIMA$+{\text{Fourier}}$
  & \textit{0.728\,{\scriptsize$\pm$0.000}}
  & 1.556\,{\scriptsize$\pm$0.000}
  & 0.470\,{\scriptsize$\pm$0.000} \\
GConvGRU$+{\text{ARIMA}}$
  & 0.779\,{\scriptsize$\pm$0.006}
  & 1.632\,{\scriptsize$\pm$0.046}
  & 0.446\,{\scriptsize$\pm$0.003} \\
GConvGRU$+{\text{ARIMA}+\text{Fourier}}$
  & \underline{0.722\,{\scriptsize$\pm$0.010}}
  & 1.478\,{\scriptsize$\pm$0.303}
  & \underline{0.438\,{\scriptsize$\pm$0.001}} \\
LSTM$+{\text{ARIMA}}$
  & 0.797\,{\scriptsize$\pm$0.005}
  & 1.573\,{\scriptsize$\pm$0.017}
  & 0.457\,{\scriptsize$\pm$0.002} \\
LSTM$+{\text{ARIMA}+\text{Fourier}}$
  & \textbf{0.713\,{\scriptsize$\pm$0.009}}
  & 1.516\,{\scriptsize$\pm$0.154}
  & \textit{0.443\,{\scriptsize$\pm$0.002}} \\
\bottomrule
\end{tabular}
\end{table*}

\paragraph{Non-traffic}
We report our results on Chickenpox (first-differenced) to connect our results to the broader literature, though we recommend future papers to report results on the undifferenced form (see Table~\ref{tab:iclr_differenced}).

Temporal models showed great strength on the non-traffic datasets, as can be seen in Table~\ref{tab:non-traffic-main}. GNNs are better, but improvements are approximately 5\%. On Chickenpox, without temporal features GConvGRU is the best, with no substantial improvement from residualization, implying that ARIMA is not providing any new information. Interestingly, the LSTM+ARIMA does better than either of the models separately. If we include Fourier features, LSTM with ARIMA residuals beats even the otherwise dominant GConvGRU. 

On the other hand, on WikiMaths, GNNs remain best with or without Fourier features, since any seasonalities seem to be captured in the input window. Residualization, as before, is not particularly helpful for the GNN models, but helps the LSTM in terms of modelling.

Finally, on PedalMe, perhaps due to its very unstable autocorrelation, temporal models do very badly, and any recovered performance need to be spatial. TDE-GNN has very good results here, in part this might be due to its explicit embedding of temporal features.

 \paragraph{Traffic}
 On the traffic datasets our re-evaluated linear models, in relative gains, show greater weakness compared to GNNs, with performance gaps between 15-35\%. This can, perhaps, be attributed to the relative homogeneity of these datasets under our measure (see Table~\ref{tab:heterogeneity}) and strong spatial and temporal signals. On traffic data, particularly PEMS03/04/07/08 flow data, residualization has a much more positive impact. Residualized GNNs consistently beat other models, including models that use day-of-week embeddings. Given the more persistent temporal signal over the input window that PEMS flow datasets have over METR-LA and PEMS-BAY (see Figure~\ref{fig:appendix_acf}), AR might be capturing it more effectively for subsequent GNNs.

 \paragraph{Is heterogeneity associated with our results?}
 Finally, we compare the GNN--temporal-baseline performance gap with the temporal strength and relative dispersion measured in Section~\ref{sub-sec:heterogeneity}. 
 The performance gap is defined by taking the difference in MSE between the best, information matched AR model, and the average MSE of GConvGRU and GraphWaveNet, to reduce the influence of a single model architecture on our results. We then fit a descriptive regression model with the performance gap as a target, and our correlation and heterogeneity statistics as features. We found that the final regression had an $\text{R}^2=0.83$, and $\text{R}_\text{adjusted}^2=0.59$. Plots of fitted metric associations, diagnostics, and further details can be found in Appendix~\ref{app:gap_associations} 
 
 This is an encouraging result and shows that heterogeneity negatively correlates with how well GNNs perform. However, cross-benchmark analysis across many models also includes many confounding variables. To isolate the effect of heterogeneity, we now move to a synthetic experiment we can control.

\begin{table*}[t] \label{tab:iclr_traffic}
\centering
\footnotesize
\setlength{\tabcolsep}{4pt}
\renewcommand{\arraystretch}{1.12}
\caption{Traffic forecasting on six datasets of BasicTS+ benchmark~\citep{shao2024exploring}. Entries are test MAE averaged over all 12 forecast steps (12 input steps; lower is better). METR-LA and PEMS-BAY use 7:1:2 train/val/test splits; PEMS03/04/07/08 use 6:2:2. \textsuperscript{$\dagger$}\,D$^2$STGNN and STID use learned day-of-week embeddings; our neural models receive time of day but no explicit day-of-week feature. Bold identifies the lowest mean among our displayed models for each dataset.}
\label{tab:traffic-main}
\begin{tabular*}{\textwidth}{@{\extracolsep{\fill}}lrrrrrr@{}}
\toprule
Model & METR-LA & PEMS-BAY & PEMS03 & PEMS04 & PEMS07 & PEMS08 \\
\midrule
\multicolumn{7}{@{}l}{\textit{Literature: BasicTS+ }} \\
\addlinespace[2pt]
DCRNN~\citep{li2018diffusion} & 3.03 & 1.59 & 15.54 & 19.66 & 21.16 & 15.23 \\
Graph WaveNet~\citep{graphwavenet} & 3.03 & 1.59 & 14.59 & 18.80 & 20.44 & 14.67 \\
MTGNN~\citep{wu2020mtgnn} & 3.05 & 1.60 & 14.85 & 19.13 & 21.01 & 15.25 \\
DGCRN~\citep{li2023dynamic}& 2.94 & 1.58 & 14.60 & 18.84 & 20.04 & 14.77 \\
D$^2$STGNN\textsuperscript{$\dagger$}~\citep{shao2022d2stgnn} & 2.88 & 1.52 & 14.63 & 18.32 & 19.49 & 14.10 \\
STNorm~\citep{deng2021st} & 3.14 & 1.58 & 15.32 & 19.21 & 20.59 & 15.39 \\
STID\textsuperscript{$\dagger$}~\citep{shao2022stid} & 3.12 & 1.56 & 15.33 & 18.35 & 19.61 & 14.21 \\
\midrule
\multicolumn{7}{@{}l}{\textit{Statistical baselines and residual models}} \\
\addlinespace[2pt]
AR(12) & 4.102 & 2.254 & 21.598 & 27.788 & 31.542 & 22.417 \\
AR(12) + Fourier & 3.747 & 2.071 & 17.083 & 22.014 & 24.523 & 17.900 \\
LSTM & 3.379 & 1.733 & 16.395 & 20.635 & 22.923 & 16.852 \\
GConvGRU + AR + Fourier & 3.216 & 1.664 & 15.158 & 19.484 & 21.058 & 15.274 \\
Graph WaveNet + AR + Fourier & \textbf{3.063} & \textbf{1.600} & \textbf{14.384} & \textbf{18.458} & \textbf{20.205} & \textbf{13.946} \\
\bottomrule
\end{tabular*}
\end{table*}

\subsection{A controlled synthetic process}
\label{sec:synthetic}

After finding that many benchmark datasets contain relatively weak spatial signals and stronger node-wise linear signals, we now investigate the role of such dataset properties by means of an easy-to-control synthetic discrete random graph process of the form:
\begin{equation}
 X_{i, t} = a_i X_{i, t-1} + \lambda \sum_{j \neq i} w_{i,j}X_{j, t-p} + 
\epsilon_{i, t}.    
\end{equation}
In other words, an AR(1) process with a graph aggregating operation, with coefficients $a_i$ encoding the temporal memory of the process, $w_{i,j}$ the node-specific spatial coupling, $\lambda$ the global graph spatial weighting, and $p$ a temporal lag in the spatial signal.
We design three synthetic test beds:  weak to strong \textit{node heterogeneity} through $\operatorname{Var}\left(\{a_i\}_{i=1}^N\right) \in [0.005,0.03]$; zero to strong \textit{spatial dispersion} through $\operatorname{Var}\left(\{w_{i,j}\}_{j=1}^N\right) \in [0,1.2]$; \textit{spatiotemporal memory} through spatial lag $p = 1,\dots,10$. Increasing the node heterogeneity and spatial dispersion thus increases the temporal and spatial heterogeneity measures $\mathcal{H}_T$ and $\mathcal{H}_G$ from Section~\ref{sub-sec:heterogeneity} respectively.

We simulate $500$ time steps on $50$ nodes and we compare ARIMA to the set of spatiotemporal models from the Chickenpox full suite~\cref{tab:iclr_differenced} on the space-time graph $X$ with a 90/10 temporal split for predicting the next step based on a 12-step input horizon. Further details regarding the synthetic experiment and the employed models are presented in Appendix \ref{sec:appendix_synth_exp} with per-model results in Table~\ref{tab:v3_axes}.
\paragraph{Results.}
\begin{figure}[t]\centering
\includegraphics[scale=0.8]{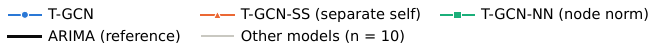}\\
\begin{subfigure}{.32\textwidth}\includegraphics[width=\linewidth]{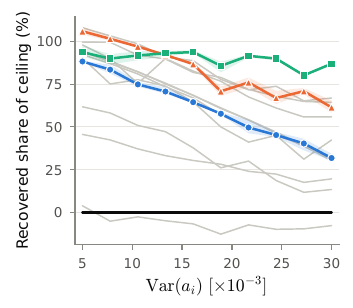}\caption{Node heterogeneity}\label{fig:synth_exp_a}\end{subfigure}\hfill
\begin{subfigure}{.32\textwidth}\includegraphics[width=\linewidth]{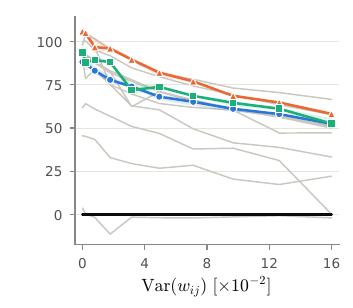}\caption{Spatial dispersion}\label{fig:synth_exp_b}\end{subfigure}\hfill
\begin{subfigure}{.32\textwidth}\includegraphics[width=\linewidth]{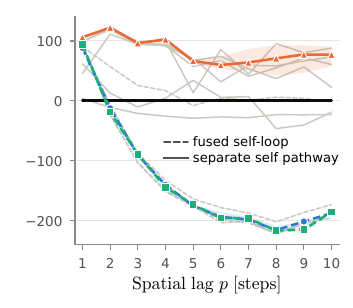}\caption{spatiotemporal memory}\label{fig:synth_exp_c}\end{subfigure}
\caption{Recovered share of spatiotemporal GNNs versus ARIMA in closed form ceiling: $100\cdot(\mathrm{MSE}_\mathrm{ARIMA}-\mathrm{MSE}_\mathrm{model})/(\mathrm{MSE}_\mathrm{own\ past}-\mathrm{Noise\ floor})$\% across three synthetic datasets varying node heterogeneity, spatial dispersion and spatiotemporal memory.}\label{fig:synth_exp}
\end{figure}

In Figure~\ref{fig:synth_exp_a} we observe a clear negative trend in model performance when increasing node-wise variance, hinting at a homophily bias similarly found in classical GNN classification tasks. As an intervention, we equip T-GCN with node-wise normalization parameters (T-GCN-NN, detailed in Appendix~\ref{sec:appendix_synth_exp}), placing it on top of the group, demonstrating the importance of node-wise specific capacity when the dataset has a certain level of temporal heterogeneity.

When looking at the spatial dispersion in Figure~\ref{fig:synth_exp_b}, we observe a similar-looking trend of increased edge-weight variance leading to worse model performance, but our node-wise intervention does not seem to be effective.
Investigating the spatiotemporal memory in Figure~\ref{fig:synth_exp_c}, we find a clear separation of models using a fused-self-loop versus models having a separate pathway, which we again confirm with a minimal architectural change, this time in form of a separate self-loop pathway within T-GCN (T-GCN-SS, detailed in Appendix~\ref{sec:appendix_synth_exp}), which significantly helps the model. Most notably, models with a fused-self loop are performing worse than ARIMA, with MPNN-LSTM the only one slightly above. This finding informs us about the relevance of self-loop treatment for modelling lagged spatiotemporal signals.

\section{Related Work}
Simple forecasting baselines challenging more complex architectures has been repeatedly demonstrated in the literature. \citet{MICHELI202285} evaluated constant predictors and node-wise ridge regression, demonstrating competitive performance on Chickenpox, PedalMe, and WikiMaths. DLinear \citep{zeng2023transformers} demonstrated the competitiveness of linear forecasters in long term-forecasting, particularly under longer than standard lookback windows. STID \citep{shao2022stid} demonstrated that an MLP equipped with learned node, time-of-day and day-of-week embeddings can achieve competitive forecasting performance without graph message passing.

This motivates careful attention to node-specific modelling and supplied temporal information. In our evaluations of temporal models, we instead try to keep them as information-matched as possible to the GNN protocol. 
Related evidence for strong classical baselines also comes from continuous-time forecasting \citep{feldman2026revisting}. Although \citet{li2018diffusion} included ARIMA, historical average, support vector regression and LSTM baselines, we revisit these baselines with more suitable specifications for traffic benchmarks. Shao et al. \citep{shao2024exploring} examine heterogeneity across datasets through temporal pattern categories and spatial sample indistinguishability, where similar historical windows are followed by distinct futures. This differs from our definition of temporal heterogeneity, since nodes can plausibly present a similar past, but still be plausibly highly homogeneous or heterogeneous.

\section{Conclusion}
In this work we analyzed datasets commonly used for benchmarking spatiotemporal regression---Chickenpox, PedalMe, WikiMaths, METR-LA, PEMS-BAY and PEMS03/04/07/08---by means of a spatiotemporal lagged dependence analysis, revealing most of these benchmark datasets are dominated by node-wise temporal information. Motivated by this, we uncovered a significant performance gap of state-of-the-art GNNs by comparing to spatially-uninformed linear ARIMA models. Furthermore, we showed how targeting the residuals of linear models based on ARIMA improved performance of GNN models.

Our analysis also revealed that the official data-loader from the PyTorch Geometric Temporal library releases first-order differenced data for Chickenpox and PedalMe, which obscures low frequency signal and weakens benchmark quality. Our findings recommend training models and running forecasts on the level (undifferenced) data.

Finally, we created a synthetic benchmark where we controlled for node-wise and edge-wise variance, and lagged spatiotemporal memory, revealing that spatiotemporal models struggled with heterogeneity and uncovered an architectural bias clearly helping to maintain lagged responses.  

\bibliography{bibliography}

\newpage
\appendix
\section{ARIMA Model Definition}
\label{sec:SARIMA}
In this section we present background details on ARIMA models, which we sometimes refer to in other sections.
Consider a time series of interest, observed up to a fixed time $t$: $\{y_\tau\}_{\tau=1}^t$,
$\tau \in \mathbb{Z}$. Additionally let $\{\epsilon_\tau\}_{\tau=1}^t$ denote the series of noise innovations, which is, as a modelling assumption, defined to be a zero-mean white 
noise process. For our purposes, we are interested
in predicting the next value $y_{t+1}$ as a linear combination of previous values of $y_t$ and previous noise innovations $\epsilon_t$. 

\begin{definition}[Backshift and differencing operators] Following notation from \cite{hyndman_forecasting_2021}, we define the backshift 
  operator $B^n$ as $B^n y_t = y_{t-n}$, and 
the $d$-th order differencing operator $\nabla^d$ as $\left(1 - B\right)^d$.
\end{definition}

The order of an ARIMA is defined by $p$ as the autoregressive (AR) order, $d$ as the differencing order, and $q$ as the moving average (MA) order.
If we define $\phi_j$ as the $j$-th autoregressive coefficient, and $\theta_j$ as the $j$-th moving average coefficient, then we can define 
the AR and MA polynomials as 
\begin{align*}
  \phi(B) = 1 - \sum_{j=1}^p \phi_j B^j, \quad
  \theta(B) = 1 + \sum_{j=1}^q \theta_j B^j.
\end{align*}


This allows us to write out the full definition of an ARIMA$(p,d,q)$ model as:
\begin{equation}
\label{eq:arima}
  \phi(B) \nabla^d y_t = \theta(B) \epsilon_t.
\end{equation}

\paragraph{Implementation Details.} By using an efficient 
implementation of AutoARIMA \citep{JSSv027i03} in Python packages StatsForecast \citep{garza2022statsforecast} and CuML \citep{raschka2020machine}, which selects the optimal order of the ARIMA based on a criterion that is minimized, we were able to scale the fitting of these even to large datasets of thousands of data points and hundreds of nodes. 

To evaluate ARIMA fairly on benchmarks which have a set default input window size, we restrict the maximum order size accordingly, such that for a benchmark
with input window of size $H$, we at most used the
$H$-lagged observation $y_{t-H}$.

      The maximal lag of the model can be found by collecting the maximal order of backshift operators in the differencing operators $\nabla$
      and the AR and MA polynomials, which follows from the additive property of the backshift operator: $\Phi(B)$ contributes a backshift operator of at most order $P$, $\phi(B)$ of at most order $p$,
      $\nabla^d$ of at most order $d$. Hence, the largest lag in the model is $p + d$.

Therefore by restricting the search space of ARIMA orders such that $p + d \leq H$, there is a fair equivalence to the windows provided by the loaders.

For the small datasets---Chickenpox, PedalMe, and WikiMaths---model selection is done by the StatsForecast implementation of AutoARIMA. Orders are chosen using a custom implementation of the AutoARIMA algorithm \citep{JSSv027i03}. For Chickenpox, we also encoded long period yearly seasonal features using Fourier Features with $K = 6$ terms, while we leveraged a SARIMA (seasonal ARIMA) for WikiMaths. This is a standard trick in time series forecasting (see \citet{hyndman_forecasting_2021}).

\section{Experimental Setup}
\label{app:experimental_setup}
\subsection{Datasets, forecast tasks, and evaluation}
\label{app:forecast_tasks}
Table~\ref{tab:datasets} summarizes the dataset tasks, size, representation, chronological split, and input and output windows.

\begin{table}[t]
\centering
\small
\caption{Dataset statistics and forecasting protocols.
$T$ counts observations before constructing forecasting windows.
$H\!\to\!K$ denotes input and forecast lengths. Three-part splits denote chronological training/validation/test
proportions.
$^{*}$Nontraffic experiments reserve the final 10\% for testing.}
\label{tab:datasets}
\setlength{\tabcolsep}{3pt}
\begin{tabular}{@{}l l r r l c c c@{}}
\toprule
\textbf{Dataset} & \textbf{Target} & \textbf{Nodes}
& \textbf{$T$} & \textbf{Freq.}
& \textbf{$H\!\to\!K$}
& \textbf{Split} \\
\midrule
Chickenpox & Cases      &    20 & 522 & Weekly
  &  $4\!\to\!1$   & $9{:}1^{*}$ \\
PedalMe    & Deliveries &    15 & 36   & Weekly
  & $4\!\to\!1$   & $9{:}1^{*}$ \\
WikiMaths  & Page views & 1,068 & 731     & Daily
   & $8\!\to\!1$   & $9{:}1^{*}$ \\
\midrule
METR-LA    & Speed      &   207 & 34,272  & 5 min
   & $12\!\to\!12$ & $7{:}1{:}2$ \\
PEMS-BAY   & Speed      &   325 & 52,116  & 5 min
   & $12\!\to\!12$ & $7{:}1{:}2$ \\
PEMS03     & Flow       &   358 & 26,208  & 5 min
    & $12\!\to\!12$ & $6{:}2{:}2$ \\
PEMS04     & Flow       &   307 & 16,992  & 5 min
     & $12\!\to\!12$ & $6{:}2{:}2$ \\
PEMS07     & Flow       &   883 & 28,224  & 5 min
   & $12\!\to\!12$ & $6{:}2{:}2$ \\
PEMS08     & Flow       &   170 & 17,856  & 5 min
   & $12\!\to\!12$ & $6{:}2{:}2$ \\
\bottomrule
\end{tabular}
\end{table}

\subsection{Neural forecasting experiments}
\label{app:neural_protocol}

The traffic experiments adopt the twelve-input,
twelve-output forecasting structure used in the BasicTS
benchmark of \citet{shao2024exploring}. Table~\ref{tab:traffic-main}
distinguishes our fitted models from published comparison
results. For our runs, model selection uses validation data
only and the test portion is reserved for reporting final
performance only. For our GNN fits, we selected 3 representative models for comparisons. We used LSTMs with hidden dimension 8, GraphWaveNet (GWN), and a GConvGRU. LSTM was fixed at hidden dimension 8 so that comparisons would not be confounded by the differing parameter counts between LSTMs and GNNs, and formed our non-linear temporal baseline. 

\subsection{Statistical baselines}
\label{app:statistical_protocol}

Statistical fits were done separately for each node, Our AR models always used the $h$ observations available within a window. Since a single linear model would underperform heavily on traffic tasks requiring a 12-step output, we instead fit one separate model for each target horizon separately. Hence, each node has 12 associated AR(12) fits done for it, with ridge regression ensuring collinearity would not break it. Ridge regression coefficients are chosen using the held-out validation set.

The ARIMA model is defined in Appendix~\ref{sec:SARIMA}. Our fits are performed using a custom AutoARIMA interpretation, which adapts work by Hyndman on automatic time series forecasting \citep{hyndman_forecasting_2021}. This allows ARIMAs to be fitted without human intervention even on large datasets. The custom auto-arima algorithm is constructed so as to constrain its search with the input window allowed by benchmark protocol, i.e. $p+d <=H$, so that the ARIMAs input windows don't accidentally cross the respective dataset protocol.

For traffic datasets, since datasets receive time-of-day information as a separate channel, we also provided the ARIMA models with Fourier features to encode a comparable measure. We also encoded time-of-day by giving each 15-minute slot in a day its own coefficient, and performing ridge-regression against the residuals from that least squares regression. In practice the 15 minute slot model and the Fourier feature variant converged to a similar ceiling for additive linear models, so our reporting focus is on the Fourier feature variant.

\subsection{Association between dependence and model gaps}
\label{app:gap_associations}

\begin{figure*}[h!]
  \centering

  \begin{subfigure}[t]{0.485\textwidth}
    \centering
    \includegraphics[width=\linewidth]{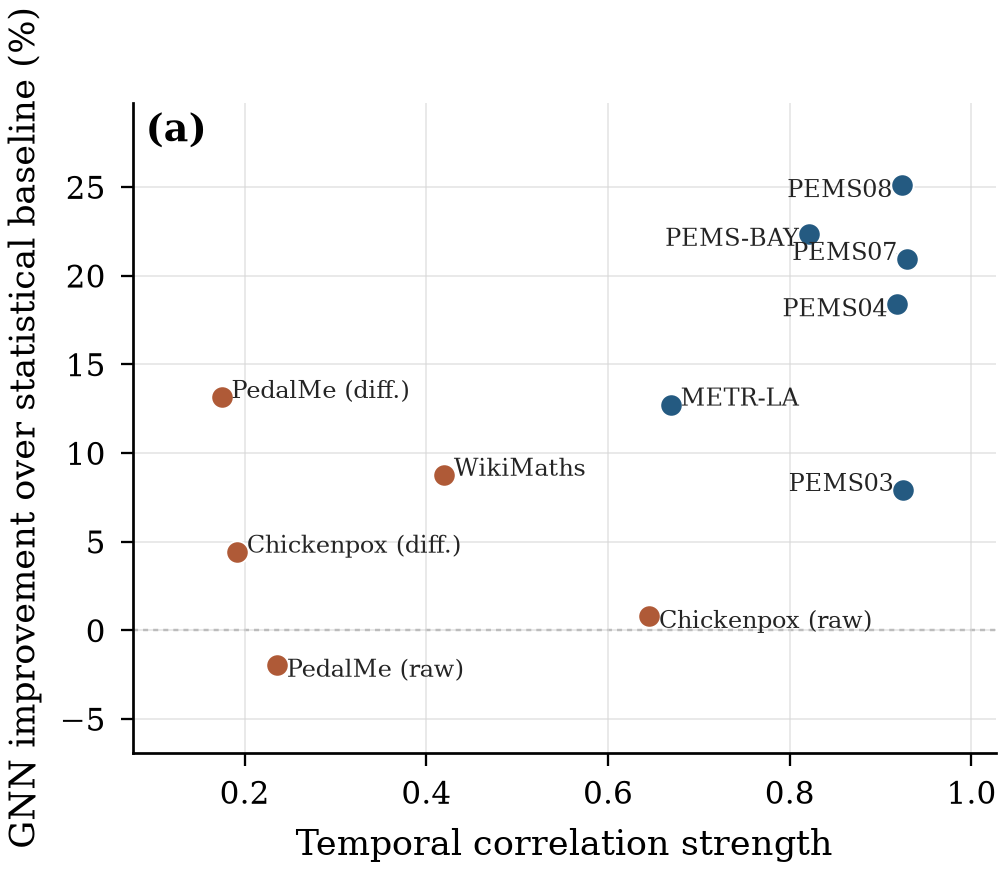}
    \phantomsubcaption\label{fig:gap-association-a}
  \end{subfigure}\hfill
  \begin{subfigure}[t]{0.485\textwidth}
    \centering
    \includegraphics[width=\linewidth]{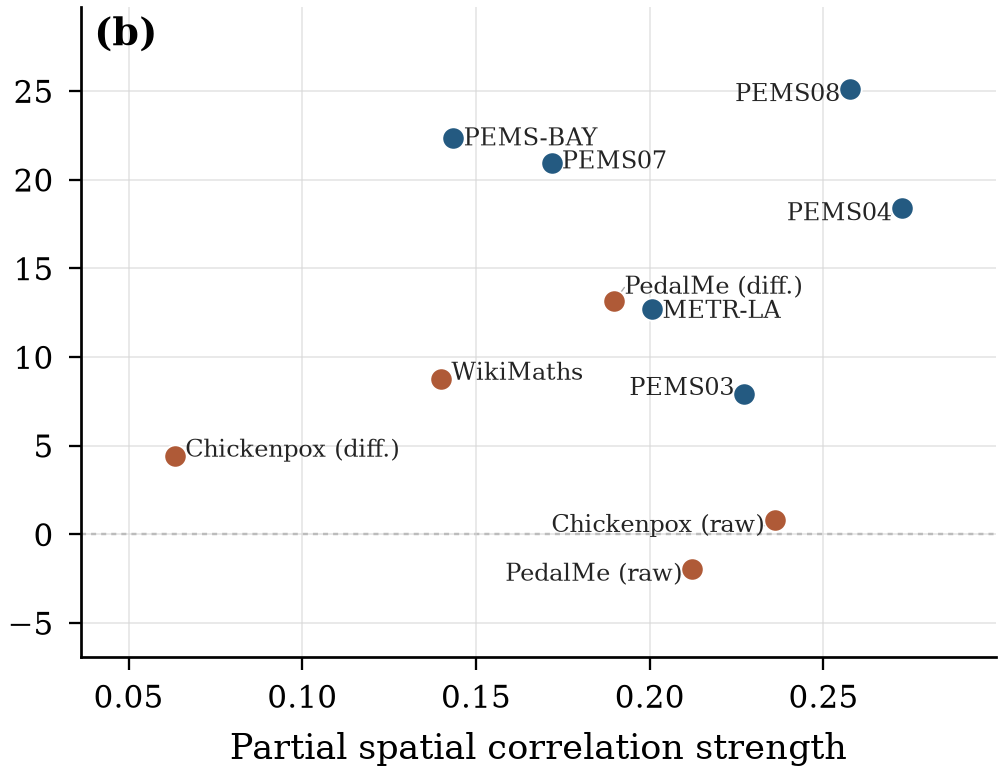}
    \phantomsubcaption\label{fig:gap-association-b}
  \end{subfigure}

  \vspace{0.35em}

  \begin{subfigure}[t]{0.485\textwidth}
    \centering
    \includegraphics[width=\linewidth]{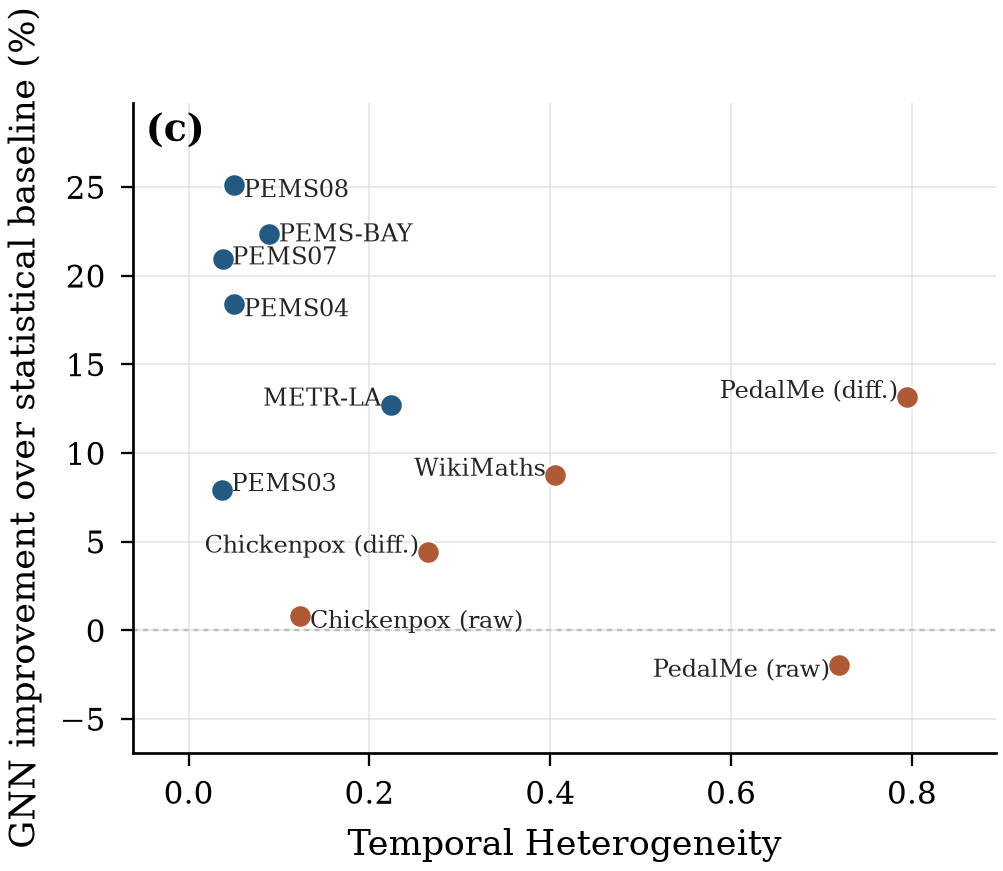}
    \phantomsubcaption\label{fig:gap-association-c}
  \end{subfigure}\hfill
  \begin{subfigure}[t]{0.485\textwidth}
    \centering
    \includegraphics[width=\linewidth]{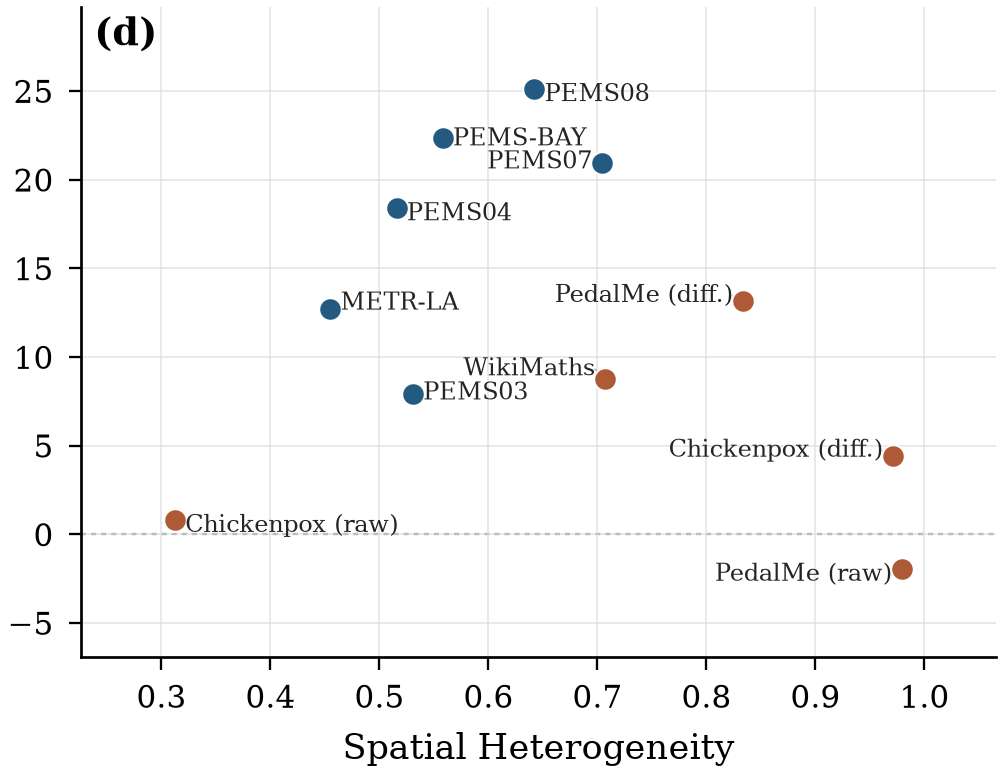}
    \phantomsubcaption\label{fig:gap-association-d}
  \end{subfigure}

  \vspace{0.35em}

  \includegraphics[width=0.2\textwidth]{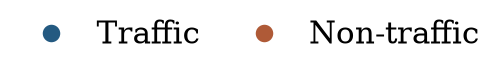}

  \caption{Association between dataset-level correlation structure and the relative MSE improvement of the mean of Graph WaveNet and direct GRU-GCN over the preselected statistical baseline. Strength is the mean absolute nodewise correlation over the input window; relative dispersion is the RMS across-node standard deviation normalized by RMS correlation magnitude. Partial spatial correlation conditions on the node's own value at the same lag. Labels identify dataset variants; colours distinguish traffic and non-traffic benchmarks.}
  \label{fig:gap-associations}
\end{figure*}

For each included dataset condition, we pair the selected
GNN and temporal-baseline errors from the same forecasting
task with the temporal-strength and heterogeneity
statistics defined in Section~\ref{sub-sec:heterogeneity}.
The signed performance gap is created by averaging the GraphWaveNet and GConvGRU performance, subtracting that from the ARIMA MSE performance, and converting that to percentages. 
We fit a regression with this gap as the target, and dataset mean temporal acf strength, mean partial-spatial acf strength, temporal heterogeneity, and partial spatial heterogeneity. In Figure~\ref{fig:gap-associations}, we plot each dataset's observed temporal baseline GNN gap against the respective features we used in the regression. The top row figures demonstrate that as partial spatial and temporal correlation increase, so does the average performance gap, whereas as dispersion increases, the performance gap decreases. There are many confounding factors at play, however, limiting the ability to generalize these findings. Hence, we attempt to isolate these relationships in the synthetic experiment of Section~\ref{sec:synthetic}.

\subsection{Experimental settings}
\label{sec:expsetup}
The large scale experiments performed on traffic datasets METRLA, PEMSBAY, PEMS03-08 were ran with a 25 CPU cluster to distribute ARIMA grid searches. The GPU used was a Nvidia-A40 46GB. The small scale experiments were ran on a home laptop, with a GTX 1060 GPU. In addition, the primary version numbers required were Python 3.12 and torch-geometric-temporal 0.56.2.

\section{Differencing Proof}\label{appendix:proof}
\begin{proof}[First order differencing is a high pass filter]
    Let the time series of interest be $Y_t$, and apply the first order differencing operator $Z_t =\nabla Y_t$. We will assume that $Y_t$ is weakly stationary and admits a spectrum $f_y(\omega)$. If we define the spectrum of $Z_t$ as $f_z(\omega)$, and $A(\omega)$ as the frequency response of a linear filter, it is a well known result that the spectrum of the output of a linear filter is $f_z(\omega) = |A(\omega)|^2$ $f_y (\omega)$. The frequency response of the operation $Z_t = \nabla Y_t$ is $A(\omega) = 1 - \exp(-i \omega)$. Therefore $|1 - \exp(-2\pi i \omega)|^2 = 2-2\cos(\omega) = 4\sin^2(\omega/2)$. Hence for small $\omega$, $f_z(\omega) \approx \omega^2 f_y(\omega)$, leading to greater dampening as $\omega$ approaches $0$. For large $\omega$ approaching the Nyquist frequency, $\pi$, then $f_z(\omega) \approx 4 f_y(\omega)$, and the filter will relatively amplify high-frequency components.
\end{proof}

\section{Differenced Chickenpox}\label{appendix:chickenpox}
Table~\ref{tab:iclr_differenced} shows the re-evaluation of Chickenpox when trained on the officially released data (standardized first difference) versus the original time series (standardized levels). 
\begin{table}[h!] 
\centering
\small
\caption{Chickenpox forecasting performance under level and
first-difference representations. MSE is measured on each
representation's standardized scale and is not directly comparable
between the two groups. Bold indicates the lowest mean test MSE
within each representation. Neural results are mean $\pm$ standard deviation over 10 seeds;
ARIMA results are deterministic.
Gap denotes the reported relative train--test gap,
$100(\mathrm{MSE}_{\mathrm{test}}-\mathrm{MSE}_{\mathrm{train}})
 /\mathrm{MSE}_{\mathrm{train}}$.
Negative values indicate lower test than training error;
they are not a ranking criterion.}
\label{tab:iclr_differenced}
\setlength{\tabcolsep}{4pt}
\renewcommand{\arraystretch}{1.08}
\begin{adjustbox}{max width=\textwidth}
\begin{tabular}{@{}l c c r c c r@{}}
\toprule
& \multicolumn{3}{c}{Standardized levels}
& \multicolumn{3}{c}{Standardized first differences} \\
\cmidrule(lr){2-4}\cmidrule(l){5-7}
Model
& Train MSE & Test MSE & Gap (\%)
& Train MSE & Test MSE & Gap (\%) \\
\midrule
ARIMA
& $0.422$ & $0.550$ & $+30.3$
& $0.639$ & $0.827$ & $+29.4$ \\

ARIMA + Fourier
& $0.388$ & $\mathbf{0.510}$ & $+31.4$
& $0.566$ & $\mathbf{0.728}$ & $+29.7$ \\

\midrule

A3T-GCN~\citep{bai2021a3t}
& $0.565 \pm 0.022$ & $0.561 \pm 0.007$ & $-0.6$
& $0.914 \pm 0.034$ & $1.069 \pm 0.014$ & $+16.9$ \\

AGCRN~\citep{bai2020agcrn}
& $0.875 \pm 0.195$ & $0.652 \pm 0.026$ & $-25.5$
& $0.849 \pm 0.041$ & $1.035 \pm 0.034$ & $+21.9$ \\

DCRNN~\citep{li2018diffusion}
& $0.484 \pm 0.022$ & $0.525 \pm 0.009$ & $+8.6$
& $0.650 \pm 0.013$ & $0.792 \pm 0.006$ & $+21.8$ \\

DyGrEncoder\citep{taheri2019predictive}
& $0.508 \pm 0.025$ & $0.528 \pm 0.009$ & $+4.0$
& $0.589 \pm 0.009$ & $0.791 \pm 0.013$ & $+34.4$ \\

EvolveGCN-H~\citep{pareja2020evolvegcn}
& $0.614 \pm 0.095$ & $0.576 \pm 0.037$ & $-6.2$
& $0.909 \pm 0.015$ & $1.039 \pm 0.018$ & $+14.3$ \\

EvolveGCN-O~\citep{pareja2020evolvegcn}
& $0.625 \pm 0.098$ & $0.548 \pm 0.009$ & $-12.3$
& $0.940 \pm 0.014$ & $1.070 \pm 0.013$ & $+13.9$ \\


GConvGRU~\citep{seo2018structured}
& $0.482 \pm 0.028$ & $0.537 \pm 0.010$ & $+11.4$
& $0.640 \pm 0.009$ & $0.799 \pm 0.004$ & $+24.7$ \\

GConvLSTM~\citep{seo2018structured}
& $0.506 \pm 0.034$ & $0.526 \pm 0.013$ & $+3.9$
& $0.591 \pm 0.026$ & $0.787 \pm 0.014$ & $+33.0$ \\


MPNN-LSTM~\citep{panagopoulos2021transfer}
& $0.508 \pm 0.058$ & $0.590 \pm 0.021$ & $+16.1$
& $0.900 \pm 0.029$ & $1.077 \pm 0.032$ & $+19.6$ \\

TDE-GNN~\citep{eliasof2024temporal}
& $0.576 \pm 0.062$ & $0.533 \pm 0.011$ & $-7.5$
& $0.830 \pm 0.006$ & $0.782 \pm 0.006$ & $-5.9$ \\

T-GCN~\citep{zhao2019t}
& $0.598 \pm 0.042$ & $0.575 \pm 0.010$ & $-3.8$
& $0.824 \pm 0.005$ & $1.041 \pm 0.012$ & $+26.3$ \\
\bottomrule
\end{tabular}
\end{adjustbox}
\end{table}

\section{Synthetic Graph Experiment}
\label{sec:appendix_synth_exp}
The node-level equation for our synthetic process is
\begin{equation}
  X_{i,t} = a_i X_{i,t-1} + \lambda\, \sum_{j \neq i} w_{i,j} X_{j,t-p} + \epsilon_{i,t},
  \label{eq:synth_node}
\end{equation}
where $a_i$ is the own-lag coefficient of node $i$, $b_i$ its spatial
coefficient, $\lambda$ a global spatial multiplier and $p \geq 1$ the lag at
which the neighbours enter. Let
\begin{align}
  X_t &= \begin{pmatrix} X_{1,t} \\ X_{2,t} \\ \vdots \\ X_{N,t} \end{pmatrix},
  & \epsilon_t &= \begin{pmatrix} \epsilon_{1,t} \\ \epsilon_{2,t} \\ \vdots \\ \epsilon_{N,t} \end{pmatrix}.
\end{align}
Define $T_A = \operatorname{diag}(a_1,\ldots,a_N)$ and the weighted adjacency matrix
$\mathbf{A} = (w_{i,j})$, with $w_{i,i} = 0$ and $\sum_j w_{i,j} = 1$ for all
$i$. In matrix form,
\begin{equation}
  X_t = T_A X_{t-1} + \lambda \mathbf{A} X_{t-p} + \epsilon_t,
  \qquad \epsilon_t \sim \mathcal{N}(0, \sigma^2 I_N).
\end{equation}
For $p = 1$ this is a VAR(1) with transition matrix
$T = T_A + \lambda \mathbf{A}$. For $p > 1$ it is a VAR($p$) with
coefficient matrices $\Phi_1 = T_A$, $\Phi_p = \lambda \mathbf{A}$ and
$\Phi_k = 0$ otherwise.
 Since the covariance of $\epsilon_t$ is positive
definite, the process is weakly stationary if and only if the spectral radius
of the companion matrix
\begin{equation}
  \mathbf{F} =
  \begin{pmatrix}
    \Phi_1 & \Phi_2 & \cdots & \Phi_{p-1} & \Phi_p \\
    I_N    & 0      & \cdots & 0          & 0      \\
    \vdots &        & \ddots &            & \vdots \\
    0      & 0      & \cdots & I_N        & 0
  \end{pmatrix}
\end{equation}
satisfies $\rho(\mathbf{F}) < 1$ \citep{tsay_analysis_2010}.

\paragraph{Node coefficients.}
The own-lag coefficients are uniformly distributed around a common mean,
\begin{equation}
  a_i \sim \mathcal{U}\!\left(\bar{a} - \sqrt{3 v_a},\ \bar{a} + \sqrt{3 v_a}\right),
\end{equation}
so that $\mathbb{E}[a_i] = \bar{a}$ and $\operatorname{Var}(a_i) = v_a$. We
use $\bar{a} = 0.5$.

\paragraph{Graph.}
The support of $\mathbf{A}$ is an undirected 3-regular graph drawn with the
configuration model. The weights follow a one-dominant pattern: each node $i$ has a dominant neighbour $j^\ast(i)$, drawn uniformly
from its three neighbours, and
\begin{equation}
  w_{i,j^\ast(i)} = \tfrac{1}{3} + u, \qquad
  w_{i,j} = \tfrac{1}{3} - \tfrac{u}{2} \;\; (j \in \mathcal{N}(i) \setminus j^\ast(i)), \qquad
  u = \tfrac{\sqrt{2}}{3}\, c .
\end{equation}
Every row sums to one and its non-zero weights have coefficient of variation
exactly $c$, so the variance of the non-zero entries of $\mathbf{A}$ is
$\operatorname{Var}(w_{ij}) = c^2 / 9$. At $c = 0$ every node averages its
neighbours uniformly; at $c = 1.2$ its weights are
$(0.90, 0.05, 0.05)$.

\paragraph{Design.}
We simulate $N = 50$ nodes for $500$ time steps with $\lambda = 0.70$ and
$\sigma = 0.1$, after a burn-in of $400$ steps from the zero state. All
configurations share one anchor $(v_a, c, p) = (0.005, 0, 1)$. We design three axis to vary: $\operatorname{Var}(a_i)$, $\operatorname{Var}(w_{ij})$ and spatial lag $p$ as reported in \cref{tab:synth_axes}. Each dataset per axis is generated from the same graph with the same assignment of nodes to coefficient quantiles and the same dominant
neighbours with the same noise sequence $\epsilon_t$. Two configurations on an
axis therefore differ only in the level of their factor, and every contrast
along an axis is a paired comparison.

\begin{table}[h]
\centering
\small
\begin{tabular}{l l l l}
\toprule
Axis & Factor & Levels & Fixed at \\
\midrule
A & node heterogeneity $v_a = \operatorname{Var}(a_i)$
  & $0.005, \ldots, 0.030$ (10, linear) & $c = 0$, $p = 1$ \\
B & weight variance $c$ ($\operatorname{Var}(w_{ij}) = c^2/9$)
  & $0, \ldots, 1.2$ (10, linear) & $v_a = 0.005$, $p = 1$ \\
C & spatial lag $p$
  & $1, \ldots, 10$ & $v_a = 0.005$, $c = 0$ \\
\bottomrule
\end{tabular}
\caption{The three axes of the synthetic benchmark.}
\label{tab:synth_axes}
\end{table}

\paragraph{Closed-form ceiling.}
Because the process is linear-Gaussian with known parameters, we compute two
reference errors exactly from its stationary autocovariances, which solve the
discrete Lyapunov equation of the companion form. The \emph{noise floor}
$\sigma^2$ is the MSE of the oracle $\mathbb{E}[X_t \mid X_{t-1}, X_{t-2},
\ldots]$, which knows the full process including the graph. The
\emph{own-past optimum} $\mathrm{MSE}_\mathrm{own}$ is the node-averaged MSE
of the best linear predictor of $X_{i,t}$ from that node's own last 12
values. Their gap $\mathrm{MSE}_\mathrm{own} - \sigma^2$ is the most that any
use of the graph can gain over a model that sees only a node's own history. We
report every model as the share of this gap it recovers over the ARIMA
reference:
\begin{equation}
  R = 100 \cdot
  \frac{\mathrm{MSE}_\mathrm{ARIMA} - \mathrm{MSE}_\mathrm{model}}
       {\mathrm{MSE}_\mathrm{own} - \sigma^2}.
\end{equation}
ARIMA sits at $R = 0$ by construction and a negative $R$ loses to it. A model
at $R = 100$ gains the full ceiling over ARIMA. Because
$\mathrm{MSE}_\mathrm{ARIMA}$ is measured on finite data, it lies somewhat
above $\mathrm{MSE}_\mathrm{own}$, which is why $R$ can exceed 100.

\paragraph{Models.}
We evaluate A3T-GCN~\citep{bai2021a3t}, AGCRN\citep{bai2020agcrn}, DCRNN~\citep{li2018diffusion}, DyGrAE~\citep{taheri2019predictive}, EvolveGCN-H~\citep{pareja2020evolvegcn}, EvolveGCN-O~\citep{pareja2020evolvegcn}, GConvGRU~\citep{seo2018structured}, GConvLSTM~\citep{seo2018structured}, MPNN-LSTM~\citep{panagopoulos2021transfer}, TDE-GNN~\citep{eliasof2024temporal}, T-GCN~\citep{zhao2019t}.

Now we elaborate on how to adapt T-GCN~\citep{zhao2019t} with a separate self pathway or node-normalisation:

Let $X_t \in \mathbb{R}^{N \times 1}$ be the input at step $t$ and
$H_t \in \mathbb{R}^{N \times d}$ the hidden state. T-GCN is a GRU whose
input transforms are graph convolutions,
\begin{align}
  Z_t &= \sigma\big([\,g_z(X_t) \,\|\, H_{t-1}\,]\, W_z + b_z\big), \\
  R_t &= \sigma\big([\,g_r(X_t) \,\|\, H_{t-1}\,]\, W_r + b_r\big), \\
  \tilde{H}_t &= \tanh\big([\,g_h(X_t) \,\|\, R_t \odot H_{t-1}\,]\, W_h + b_h\big), \\
  H_t &= Z_t \odot H_{t-1} + (1 - Z_t) \odot \tilde{H}_t,
\end{align}
with $W_\cdot \in \mathbb{R}^{2d \times d}$ shared across nodes, followed by
a shared readout of the last state, $\hat{X}_{i,t+1} = w^\top
\operatorname{ReLU}(h_{i,t}) + c$, where $h_{i,t}$ is row $i$ of $H_t$.
In T-GCN each gate's convolution is a GCN layer with self-loops,
\begin{equation}
  g_k(X_t) = \hat{\mathbf{A}}\, X_t\, \Theta_k + \mathbf{1} b_k^\top,
  \qquad
  \hat{\mathbf{A}} = \tilde{D}^{-1/2} (\mathbf{A}_\mathrm{s} + I)\, \tilde{D}^{-1/2},
  \qquad k \in \{z, r, h\},
  \label{eq:tgcn_fused}
\end{equation}
so node $i$'s own input and its neighbours' enter through the same
$\Theta_k \in \mathbb{R}^{1 \times d}$, weighted by the non-negative
$\hat{A}_{ii}$ and $\hat{A}_{ij}$: the two coefficients cannot differ in
sign or be learned independently.

\emph{T-GCN-SS} (separate self pathway) replaces \eqref{eq:tgcn_fused} in
all three gates by
\begin{equation}
  g_k^{\mathrm{SS}}(X_t)
  = X_t\, \Theta_k^{\mathrm{self}} + \mathbf{1} b_k^\top
  + \hat{\mathbf{A}}_0\, X_t\, \Theta_k^{\mathrm{nbr}},
  \qquad
  \hat{\mathbf{A}}_0 = D^{-1/2} \mathbf{A}_\mathrm{s} D^{-1/2},
\end{equation}
where the neighbour aggregation carries no self-loop and
$\Theta_k^{\mathrm{self}}, \Theta_k^{\mathrm{nbr}} \in \mathbb{R}^{1 \times d}$
are separate. Everything else is unchanged. This adds $3d$ shared parameters
($96$ at $d = 32$).

\emph{T-GCN-NN} (node normalisation) keeps the recurrence of T-GCN and
changes only the readout. The last hidden state of each node is standardised
across its $d$ channels and then scaled and shifted by one scalar pair per
node:
\begin{equation}
  \bar{h}_{i,t} = \frac{h_{i,t} - \mu(h_{i,t})}{\sigma(h_{i,t})},
  \qquad
  \hat{X}_{i,t+1} = w^\top \operatorname{ReLU}\big(\gamma_i\, \bar{h}_{i,t} + \beta_i\big) + c,
\end{equation}
where $\mu(\cdot)$ and $\sigma(\cdot)$ are the mean and standard deviation
over the channels of $h_{i,t}$ and $\gamma_i, \beta_i \in \mathbb{R}$ are
initialised to $1$ and $0$. This adds $2N = 100$ node-specific parameters.

\paragraph{Training and evaluation.}
All models forecast one step ahead based on an input window of 12 steps. Samples are indexed by target time and
split temporally into train $[12, 350)$, validation $[350, 400)$ and test
$[400, 500)$. The first 12 targets are dropped so that every model is scored
on identical targets. The series is standardised with the mean and standard
deviation of the training slice, and all MSEs are reported on the original
scale. The optimizer employed is Adam~\citep{kingma_adam_2015}.

Hyper-parameters are tuned separately for every model on every dataset and listed in \cref{tab:synth_hparams}. For
each model we draw 24 configurations at random from its search space (or
enumerate the space when it is smaller) and train each with Adam on the MSE
for at most 200 epochs. Training uses early stopping on validation MSE
(patience 25), halves the learning rate on a plateau (patience 10) and clips
gradients at norm 5. We keep the configuration with the lowest validation
MSE and refit it with three training seeds on the same dataset; reported
numbers are the mean test MSE over these seeds. 

\begin{table}[h]
\centering
\small
\setlength{\tabcolsep}{5pt}
\begin{tabular}{@{}l l l@{}}
\toprule
Model & Hyper-parameter & Values \\
\midrule
\multicolumn{3}{@{}l}{\textit{Base Values}} \\
 & Learning rate & $\{3\cdot10^{-4},\ 10^{-3},\ 3\cdot10^{-3},\ 10^{-2},\ 3\cdot10^{-2}\}$ \\
 & Weight decay  & $\{0,\ 10^{-5},\ 10^{-4},\ 10^{-3}\}$ \\
 & Hidden size   & $\{16,\ 32,\ 64\}$ \\
 & Batch size    & $\{32,\ 64,\ 128\}$ \\
 & Dropout       & $\{0,\ 0.1,\ 0.3\}$ \\
 & Input window  & 12 (fixed) \\
\midrule
\multicolumn{3}{@{}l}{\textit{Model-specific additions}} \\
GConvGRU/LSTM, DCRNN & Filter order $K$ & $\{1,\ 2,\ 3\}$ \\
\addlinespace[2pt]
DyGrAE & Graph-conv layers $K$ & $\{1,\ 2,\ 3\}$ \\
       & Conv channels         & $\{4,\ 8,\ 16\}$ \\
\addlinespace[2pt]
AGCRN  & Filter order $K$      & $\{1,\ 2,\ 3\}$ \\
       & Node-embedding dim.   & $\{2,\ 4,\ 8\}$ \\
\midrule
\multicolumn{3}{@{}l}{\textit{Custom search space}} \\
TDE-GNN & Learning rate (I/O, temporal, spatial) & 7 log-spaced in $[10^{-4},\ 10^{-1}]$ \\
        & Weight decay (I/O, temporal, spatial)  & $\{0,\ 10^{-4},\ 10^{-3},\ 5\cdot10^{-3},\ 10^{-2}\}$ \\
        & Dropout (I/O, hidden)                  & $\{0,\ 0.1,\ 0.3,\ 0.5,\ 0.7,\ 0.9\}$ \\
        & Hidden size                            & $\{8,\ 16,\ 32,\ 64,\ 128,\ 256\}$ \\
        & Step size $h$                          & $\{10^{-3},\ 10^{-2},\ 0.1,\ 0.3,\ 0.5,\ 1\}$ \\
        & Batch norm                             & $\{$on, off$\}$ \\
\addlinespace[2pt]
ARIMA   & Order $(p', d, q)$    & $p' \in \{1,2,3\},\ d = 0,\ q \in \{0,1,2\}$, w/wo mean \\
        &                       & $p' \in \{0,1,2\},\ d = 1,\ q \in \{0,1\}$ \\
\bottomrule
\end{tabular}
\caption{Hyper-parameter search spaces. T-GCN, A3T-GCN,
EGCN-O, EGCN-H and MPNN-LSTM use the shared space only. For every model and dataset, 24 configurations are sampled
uniformly (the 24 ARIMA orders are enumerated) and the one with the lowest
validation MSE is refit with three training seeds.}
\label{tab:synth_hparams}
\end{table}

\paragraph{Results.}
Table \ref{tab:v3_axes} shows the model-by-model readings for the benchmark, aggregating the ten configurations per axis into its min, mean, and max properties.
\begin{table}[h]
\centering
\caption{Recovered share of the closed-form graph ceiling (\%) along the three axes of the synthetic benchmark $100\cdot(\mathrm{MSE}_\mathrm{ARIMA}-\mathrm{MSE}_\mathrm{model})/(\mathrm{MSE}_\mathrm{own\ past}-\sigma^2)$. Type: F = fused into one normalized $A+I$ operator, S = separate pathway. Values obtained from three seeds (0,1,2) and input-output horizon of 12$\rightarrow$1. Bold: best per column.}
\label{tab:v3_axes}
\small
\setlength{\tabcolsep}{4pt}
\begin{tabular}{l c rrr rrr rrr}
\toprule
Model & Type & \multicolumn{3}{c}{$\mathrm{Var}(a_i)$} & \multicolumn{3}{c}{$\mathrm{Var}(w_{ij})$} & \multicolumn{3}{c}{spatial lag $p$} \\
\cmidrule(lr){3-5} \cmidrule(lr){6-8} \cmidrule(lr){9-11}
 & & min & mean & max & min & mean & max & min & mean & max \\
\midrule
A3T-GCN & F & $30.9$ & $62.3$ & $92.3$ & $51.0$ & $73.6$ & $92.3$ & $-202.1$ & $-122.7$ & $92.3$ \\
AGCRN & S & $-12.8$ & $-6.3$ & $3.8$ & $-11.3$ & $-1.8$ & $3.8$ & $-29.2$ & $-20.8$ & $3.8$ \\
DCRNN & S & $55.8$ & $75.3$ & $98.0$ & $33.3$ & $66.1$ & $104.0$ & $13.4$ & $82.4$ & $122.0$ \\
DyGrAE & S & $65.2$ & $84.5$ & $\mathbf{108.0}$ & $\mathbf{66.6}$ & $\mathbf{86.9}$ & $\mathbf{108.0}$ & $22.5$ & $72.7$ & $118.8$ \\
DynGESN & S & $11.6$ & $24.9$ & $37.2$ & $19.2$ & $28.9$ & $36.3$ & $25.6$ & $56.8$ & $118.9$ \\
EGCN-H & F & $30.2$ & $63.9$ & $94.1$ & $51.1$ & $74.2$ & $94.1$ & $-221.5$ & $-138.6$ & $94.1$ \\
EGCN-O & F & $31.2$ & $64.0$ & $94.6$ & $50.0$ & $72.5$ & $94.6$ & $-222.2$ & $-138.0$ & $94.6$ \\
GC-LSTM & S & $17.5$ & $30.0$ & $45.6$ & $17.4$ & $31.2$ & $45.6$ & $40.9$ & $69.5$ & $111.3$ \\
GCRN-GRU & S & $64.4$ & $81.1$ & $99.6$ & $57.9$ & $81.7$ & $100.4$ & $31.6$ & $76.8$ & $118.7$ \\
HMM4G & S & $23.8$ & $50.9$ & $82.7$ & $23.7$ & $56.0$ & $82.7$ & $-35.3$ & $-4.3$ & $82.7$ \\
MPNN-LSTM & F & $31.1$ & $59.1$ & $90.2$ & $47.2$ & $67.9$ & $90.2$ & $-8.0$ & $19.6$ & $90.2$ \\
T-GCN & F & $31.8$ & $60.6$ & $88.3$ & $52.5$ & $71.6$ & $88.3$ & $-216.1$ & $-132.9$ & $88.3$ \\
T-GCN-NN & F & $\mathbf{80.1}$ & $\mathbf{89.7}$ & $93.8$ & $52.6$ & $75.2$ & $93.7$ & $-216.8$ & $-134.3$ & $93.7$ \\
T-GCN-SS & S & $61.4$ & $82.9$ & $105.9$ & $58.2$ & $84.4$ & $105.9$ & $\mathbf{59.7}$ & $\mathbf{84.2}$ & $\mathbf{122.2}$ \\
TDE-GNN & S & $11.7$ & $35.5$ & $61.7$ & $0.3$ & $45.0$ & $64.1$ & $-46.5$ & $0.9$ & $61.7$ \\
\bottomrule
\end{tabular}
\end{table}

\newpage
\section{Full ACF and dispersion panel}\label{appendix:statistics}
\begin{figure*}[h!]
\centering
\begin{subfigure}[t]{0.32\textwidth}
\centering
\includegraphics[width=\linewidth]{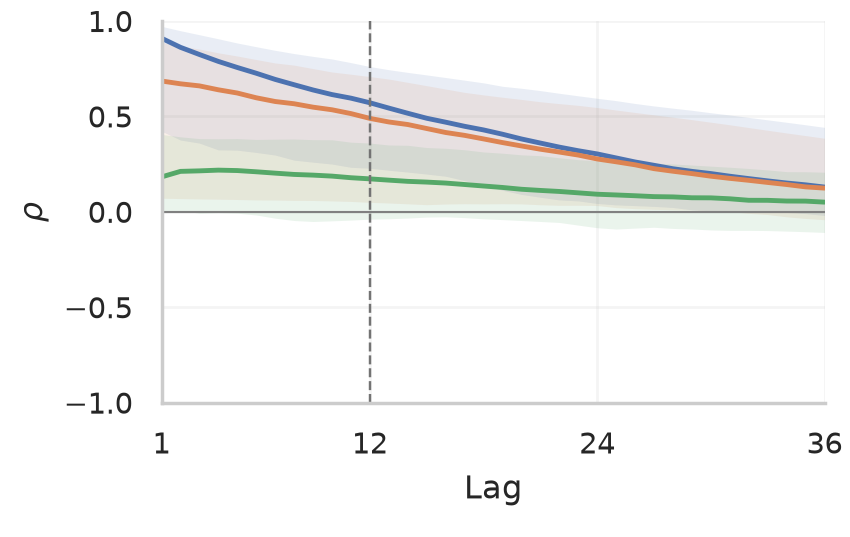}
\caption{METR-LA}
\end{subfigure}
\hfill
\begin{subfigure}[t]{0.32\textwidth}
\centering
\includegraphics[width=\linewidth]{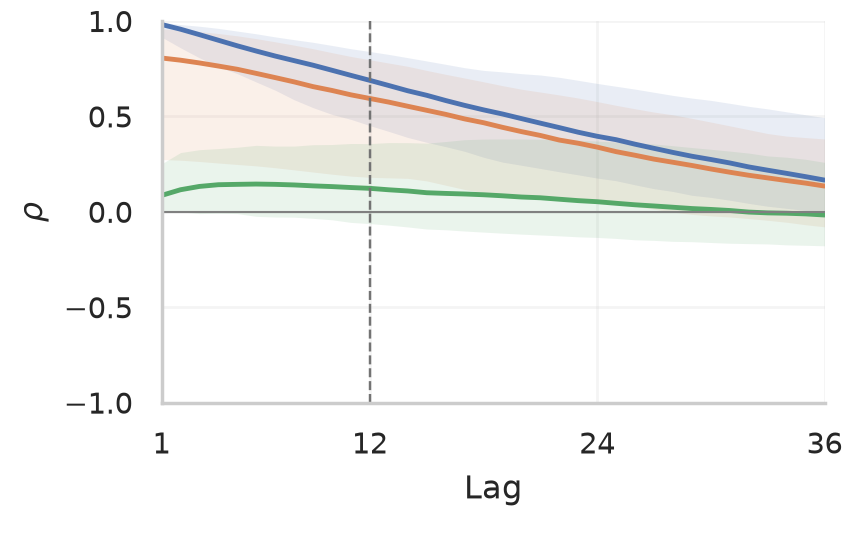}
\caption{PEMS-BAY}
\end{subfigure}
\hfill
\begin{subfigure}[t]{0.32\textwidth}
\centering
\includegraphics[width=\linewidth]{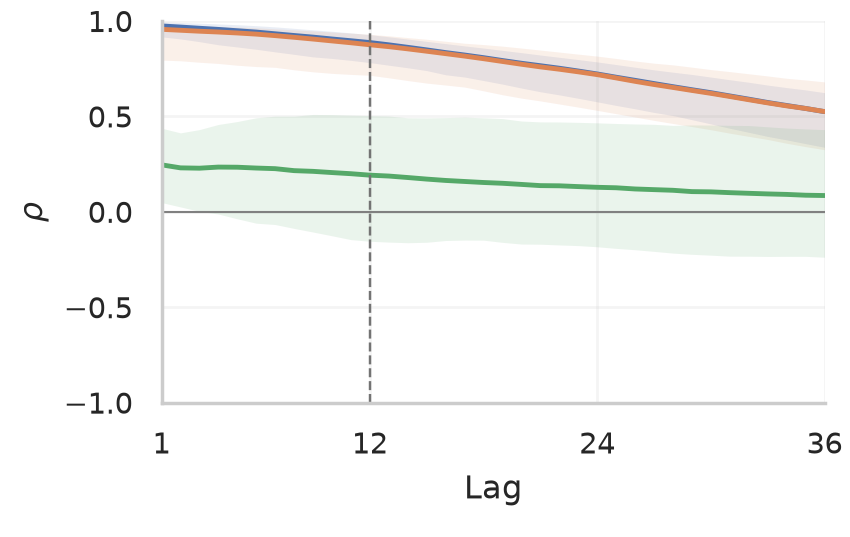}
\caption{PEMS03}
\end{subfigure}
\par\smallskip
\begin{subfigure}[t]{0.32\textwidth}
\centering
\includegraphics[width=\linewidth]{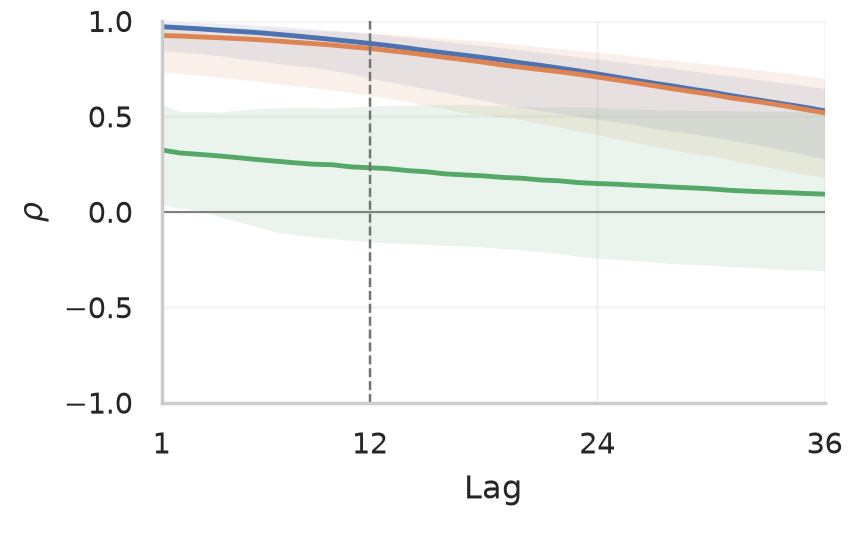}
\caption{PEMS04}
\end{subfigure}
\hfill
\begin{subfigure}[t]{0.32\textwidth}
\centering
\includegraphics[width=\linewidth]{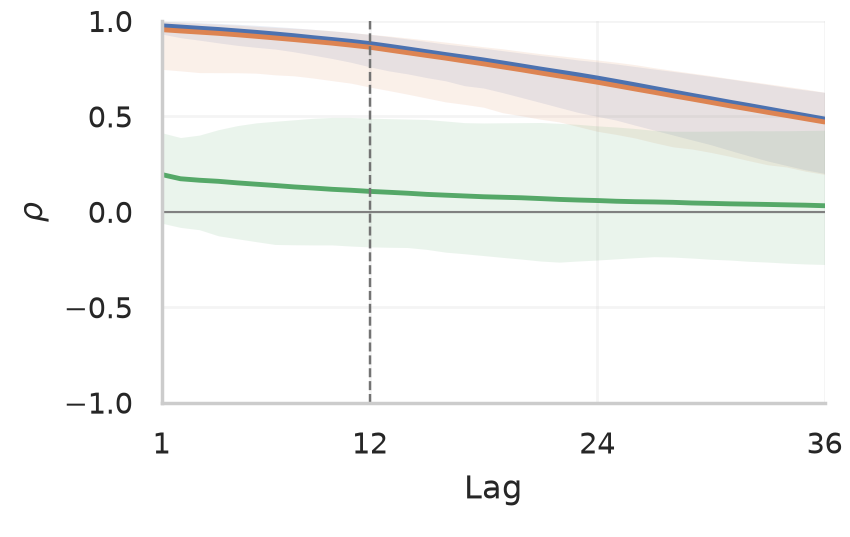}
\caption{PEMS07}
\end{subfigure}
\hfill
\begin{subfigure}[t]{0.32\textwidth}
\centering
\includegraphics[width=\linewidth]{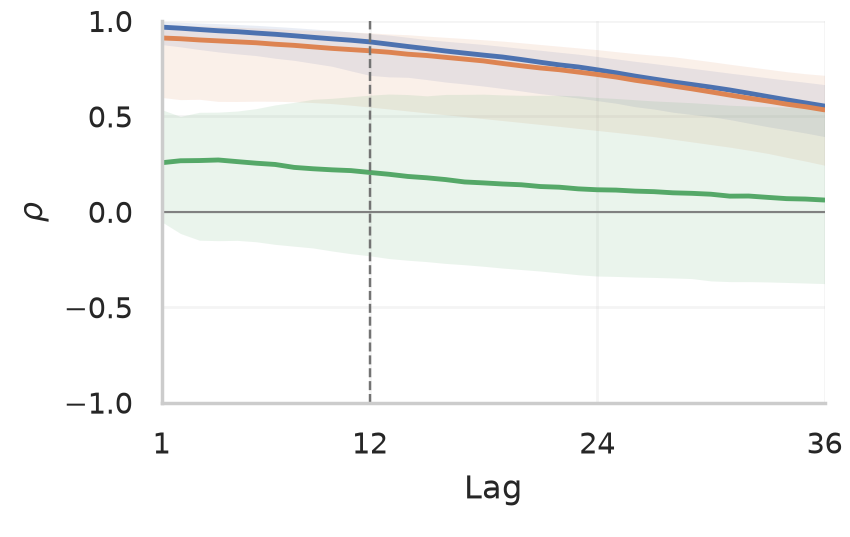}
\caption{PEMS08}
\end{subfigure}
\par\smallskip
\begin{subfigure}[t]{0.32\textwidth}
\centering
\includegraphics[width=\linewidth]{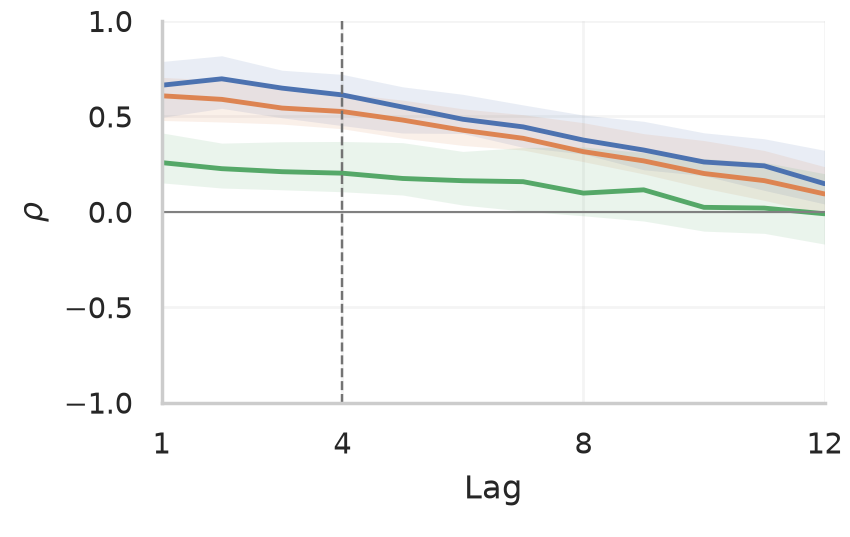}
\caption{Chickenpox (levels)}
\end{subfigure}
\hfill
\begin{subfigure}[t]{0.32\textwidth}
\centering
\includegraphics[width=\linewidth]{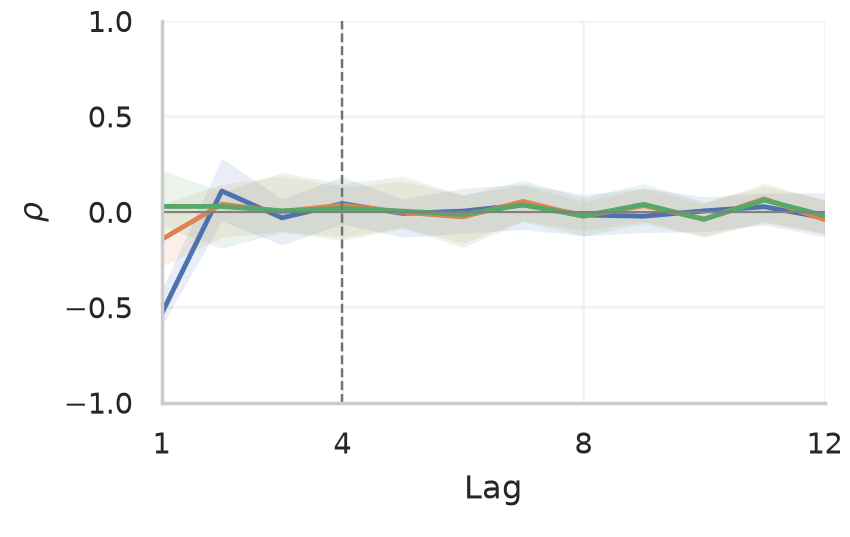}
\caption{Chickenpox (differences)}
\end{subfigure}
\hfill
\begin{subfigure}[t]{0.32\textwidth}
\centering
\includegraphics[width=\linewidth]{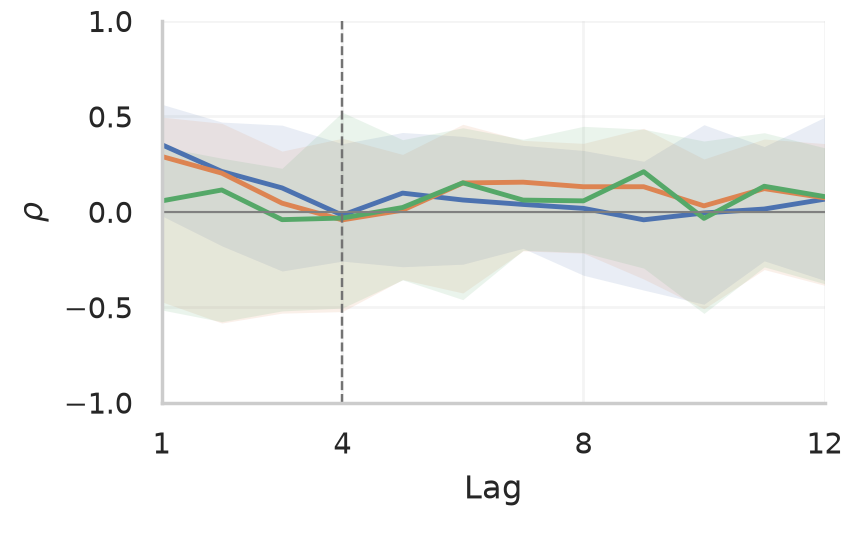}
\caption{PedalMe (levels)}
\end{subfigure}
\par\smallskip
\begin{subfigure}[t]{0.32\textwidth}
\centering
\includegraphics[width=\linewidth]{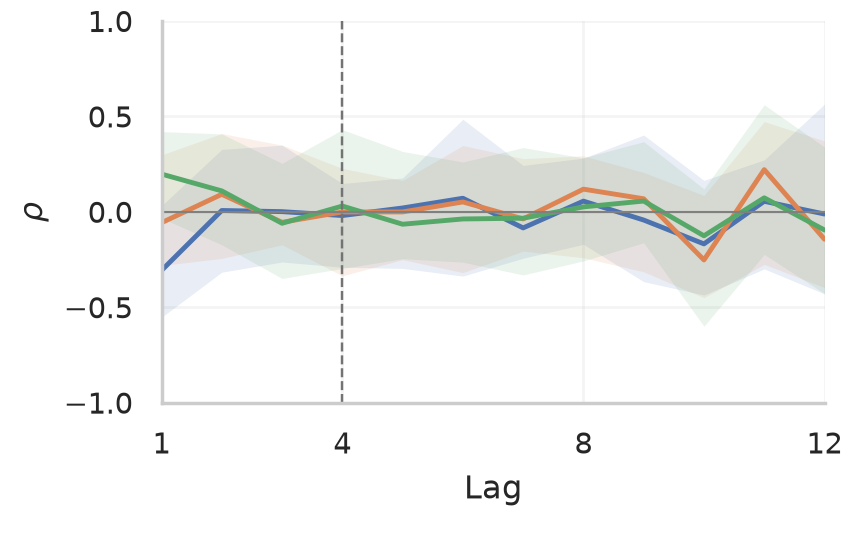}
\caption{PedalMe (differences)}
\end{subfigure}
\hfill
\begin{subfigure}[t]{0.32\textwidth}
\centering
\includegraphics[width=\linewidth]{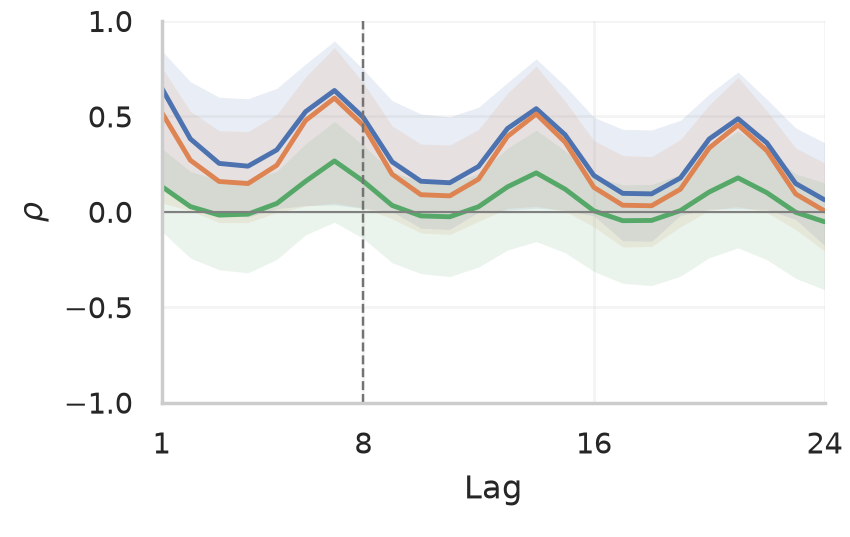}
\caption{WikiMaths}
\end{subfigure}
\hfill
\par\smallskip
\includegraphics[width=0.72\textwidth]{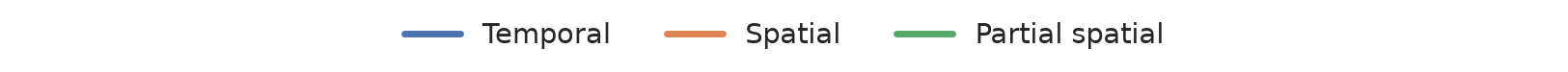}
\caption{Nodewise temporal, spatial and partial spatial correlations. Lines show medians; shading spans the central 95\% across nodes, not confidence intervals. Lags extend to $3H$; the dashed vertical line marks $H$. A common set of nodes with finite estimates for all three statistics at every plotted lag is used.}
\label{fig:appendix_acf}
\end{figure*}

\begin{figure*}[h!]
\centering
\begin{subfigure}[t]{0.32\textwidth}
\centering
\includegraphics[width=\linewidth]{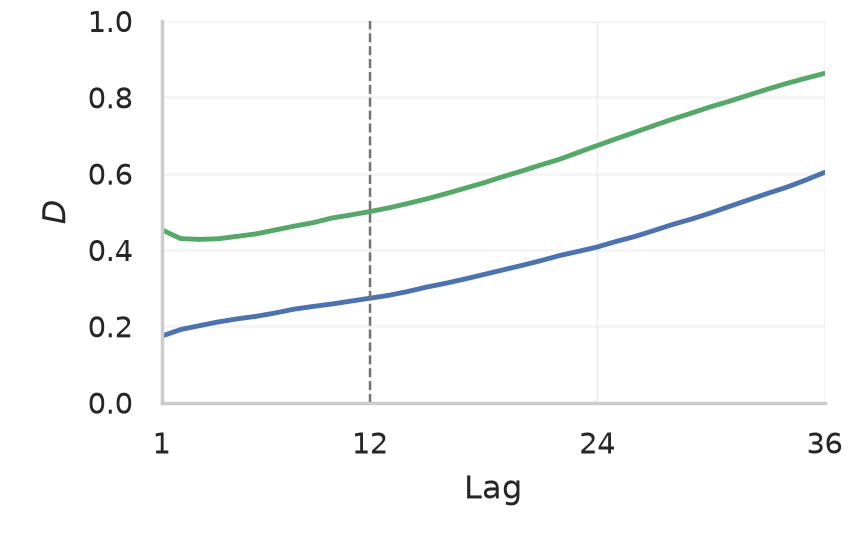}
\caption{METR-LA}
\end{subfigure}
\hfill
\begin{subfigure}[t]{0.32\textwidth}
\centering
\includegraphics[width=\linewidth]{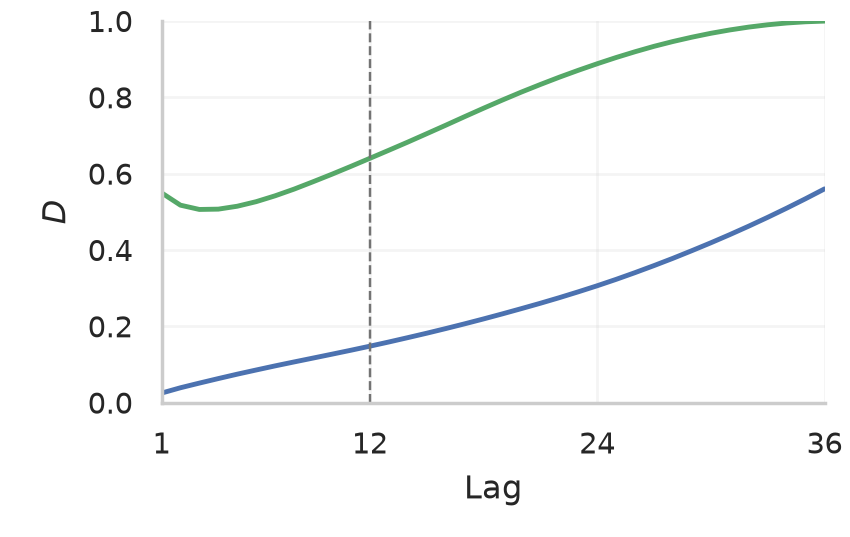}
\caption{PEMS-BAY}
\end{subfigure}
\hfill
\begin{subfigure}[t]{0.32\textwidth}
\centering
\includegraphics[width=\linewidth]{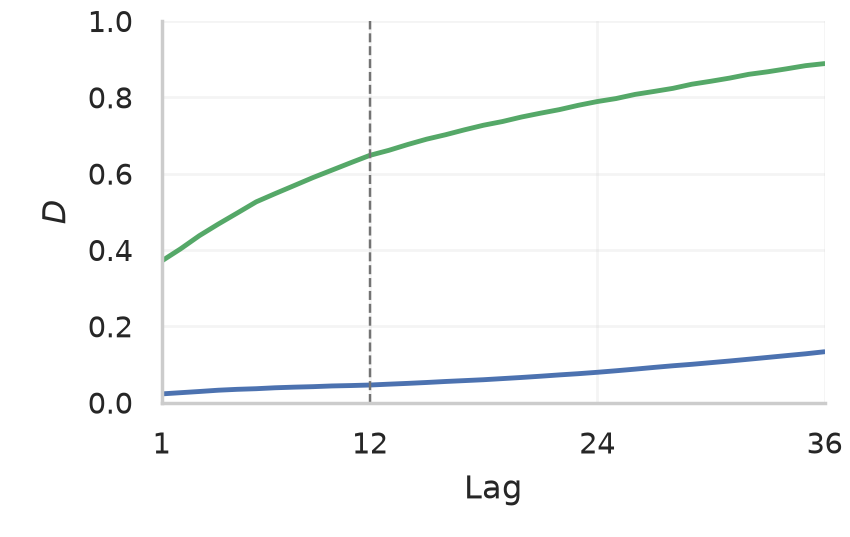}
\caption{PEMS03}
\end{subfigure}
\par\smallskip
\begin{subfigure}[t]{0.32\textwidth}
\centering
\includegraphics[width=\linewidth]{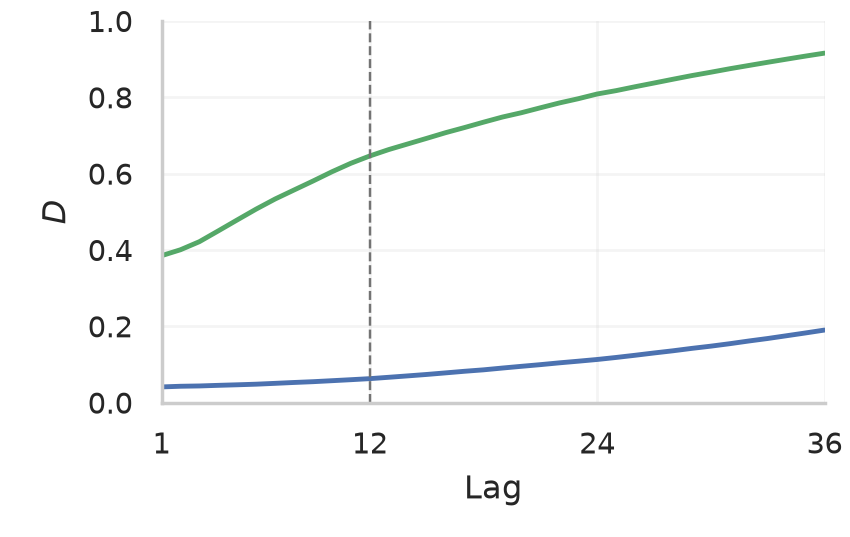}
\caption{PEMS04}
\end{subfigure}
\hfill
\begin{subfigure}[t]{0.32\textwidth}
\centering
\includegraphics[width=\linewidth]{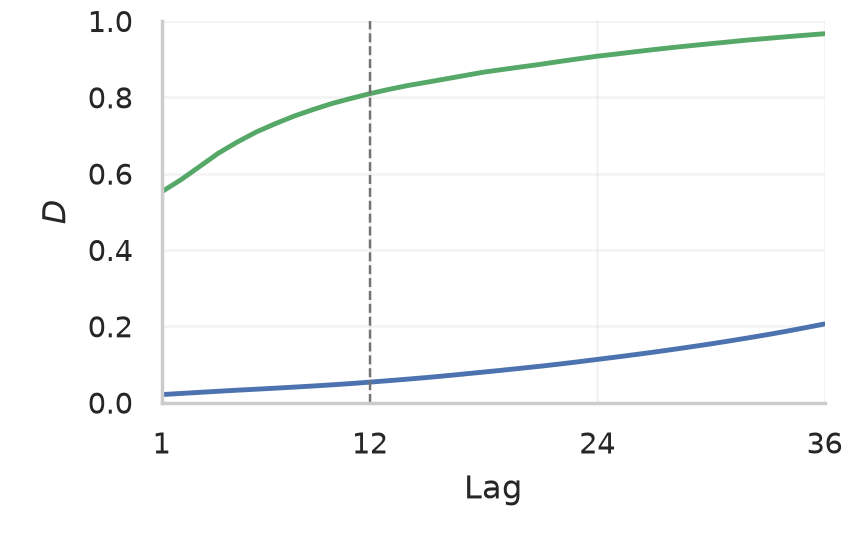}
\caption{PEMS07}
\end{subfigure}
\hfill
\begin{subfigure}[t]{0.32\textwidth}
\centering
\includegraphics[width=\linewidth]{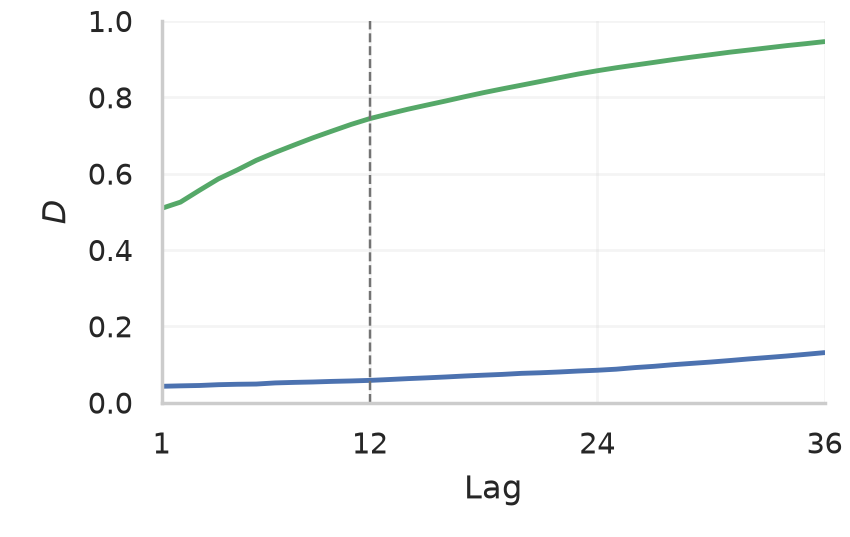}
\caption{PEMS08}
\end{subfigure}
\par\smallskip
\begin{subfigure}[t]{0.32\textwidth}
\centering
\includegraphics[width=\linewidth]{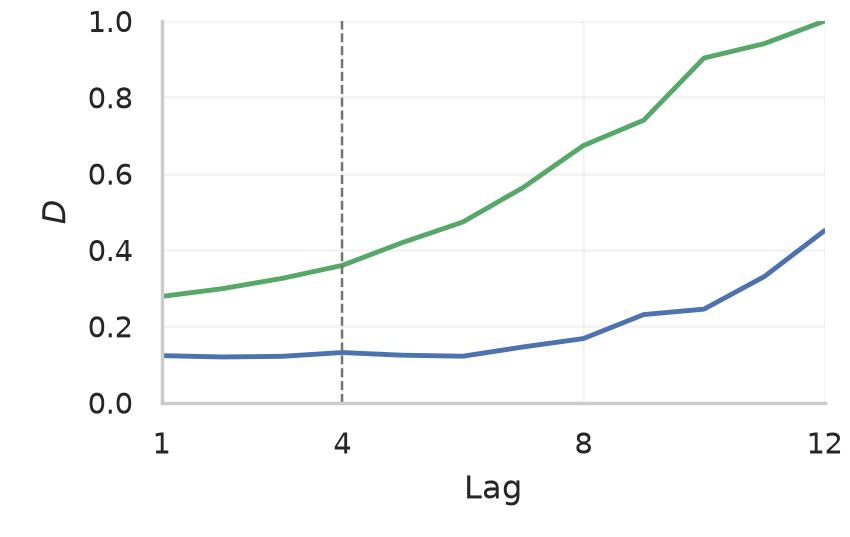}
\caption{Chickenpox (levels)}
\end{subfigure}
\hfill
\begin{subfigure}[t]{0.32\textwidth}
\centering
\includegraphics[width=\linewidth]{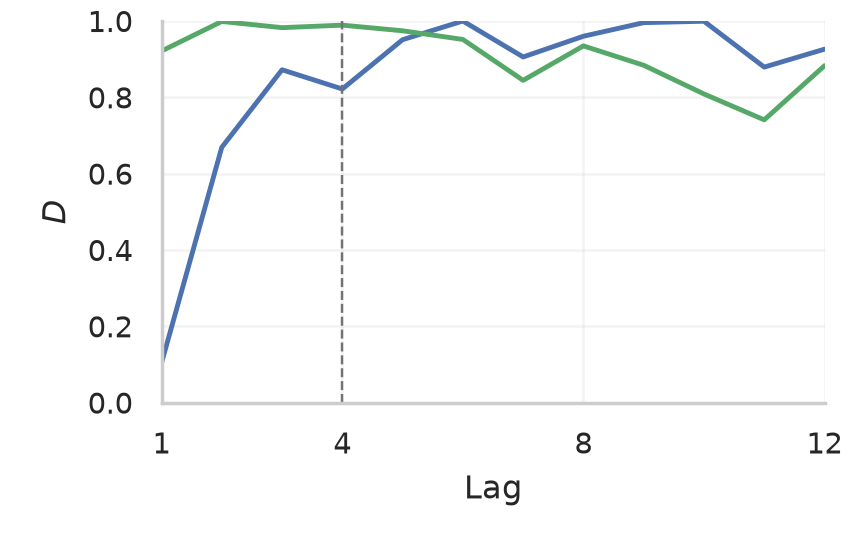}
\caption{Chickenpox (differences)}
\end{subfigure}
\hfill
\begin{subfigure}[t]{0.32\textwidth}
\centering
\includegraphics[width=\linewidth]{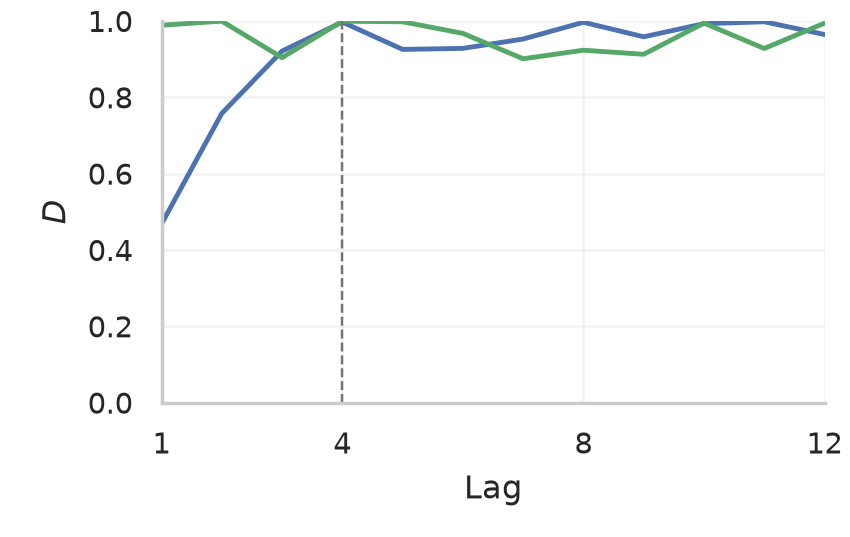}
\caption{PedalMe (levels)}
\end{subfigure}
\par\smallskip
\begin{subfigure}[t]{0.32\textwidth}
\centering
\includegraphics[width=\linewidth]{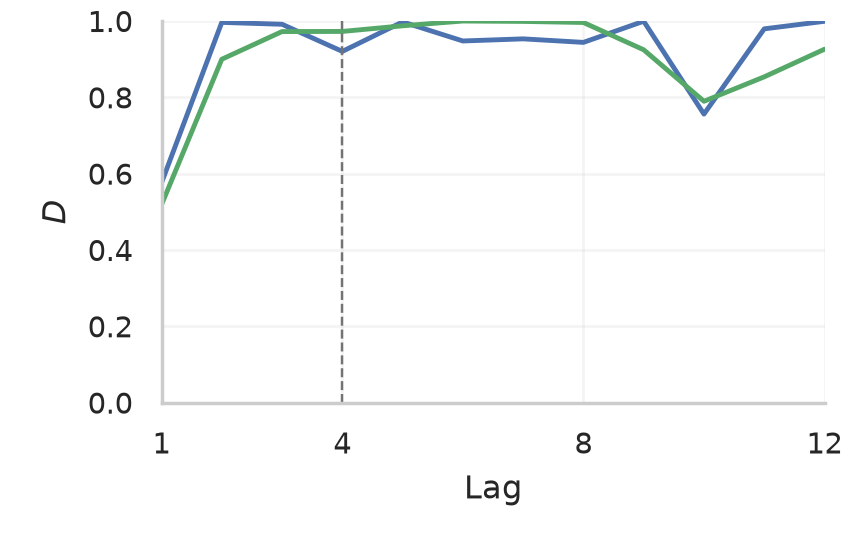}
\caption{PedalMe (differences)}
\end{subfigure}
\hfill
\begin{subfigure}[t]{0.32\textwidth}
\centering
\includegraphics[width=\linewidth]{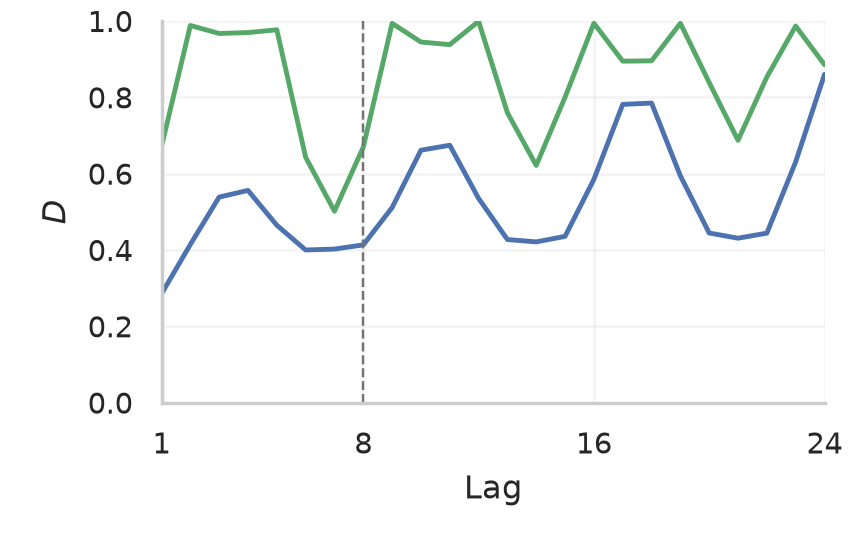}
\caption{WikiMaths}
\end{subfigure}
\hfill
\par\smallskip
\includegraphics[width=0.72\textwidth]{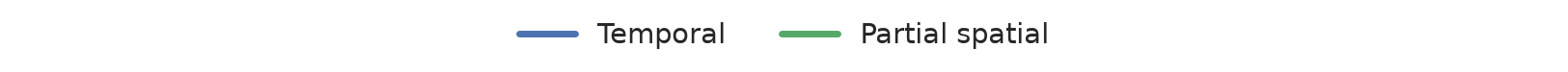}
\caption{Lagwise relative dispersion of temporal and partial spatial node-wise correlations, $\mathcal{H}=\mathrm{SD}/\mathrm{RMS}$. Note that high relative dispersion can occur even when correlations are weak. Lags extend to $3H$; the dashed vertical line marks $H$.}
\label{fig:appendix_dispersion}
\end{figure*}

In Figure \ref{fig:appendix_acf} we present the full set of node-wise temporal, spatial and partial spatial correlations. Figure \ref{fig:appendix_dispersion} shows the full set of lag-wise relative dispersion of temporal and partial spatial node-wise correlations.
Table \ref{tab:heterogeneity} complements this with quantitative summaries per dataset's standard horizon $H$.

\begin{table}[t]
\centering
\caption{Temporal and partial spatial heterogeneity
within each dataset's input window.}
\label{tab:heterogeneity}
\small
\setlength{\tabcolsep}{5pt}
\begin{tabular}{@{}lrr@{}}
\toprule
Dataset & Temporal $\mathcal{H}_T$ & Partial spatial $\mathcal{H}_G$ \\
\midrule
METR-LA & 0.223 & 0.455 \\
PEMS-BAY & 0.089 & 0.558 \\
PEMS03 & 0.037 & 0.531 \\
PEMS04 & 0.051 & 0.516 \\
PEMS07 & 0.038 & 0.705 \\
PEMS08 & 0.050 & 0.642 \\
\addlinespace[2pt]
Chickenpox (levels) & 0.124 & 0.313 \\
Chickenpox (differences) & 0.265 & 0.971 \\
\addlinespace[2pt]
PedalMe (levels) & 0.720 & 0.979 \\
PedalMe (differences) & 0.795 & 0.834 \\
\addlinespace[2pt]
WikiMaths & 0.405 & 0.707 \\
\bottomrule
\end{tabular}
\end{table}

\end{document}